\documentclass[sigconf]{acmart}

\usepackage{fancyhdr}

\usepackage{graphicx}
\usepackage{subfigure}
\usepackage{booktabs} 
\usepackage{caption}
\usepackage{xcolor}
\usepackage{stfloats}
\usepackage{algorithmic}
\usepackage{algorithm}
\usepackage{multirow}

\usepackage{hyperref}
\usepackage{newtxmath}
\usepackage{amsmath}

\definecolor{Note_color}{rgb}{0.0, 0.0, 1.0}

\AtBeginDocument{%
  \providecommand\BibTeX{{%
    \normalfont B\kern-0.5em{\scshape i\kern-0.25em b}\kern-0.8em\TeX}}}

\copyrightyear{2021}
\acmYear{2021}
\setcopyright{acmcopyright}\acmConference[MICRO '21]{MICRO-54: 54th Annual IEEE/ACM International Symposium on Microarchitecture}{October 18--22, 2021}{Virtual Event, Greece}
\acmBooktitle{MICRO-54: 54th Annual IEEE/ACM International Symposium on Microarchitecture (MICRO '21), October 18--22, 2021, Virtual Event, Greece}
\acmPrice{15.00}
\acmDOI{10.1145/3466752.3480082}
\acmISBN{978-1-4503-8557-2/21/10}

\begin{document}

\title{2-in-1 Accelerator: Enabling Random Precision Switch for Winning Both Adversarial Robustness and Efficiency}

\author{Yonggan Fu}
\affiliation{%
  \institution{Rice University}
  \country{}
}
\email{yf22@rice.edu}

\author{Yang Zhao}
\affiliation{%
  \institution{Rice University}
  \country{}
}
\email{zy34@rice.edu}

\author{Qixuan Yu}
\affiliation{%
  \institution{Rice University}
  \country{}
}
\email{Maki.Yu@rice.edu }

\author{Chaojian Li}
\affiliation{%
  \institution{Rice University}
  \country{}
}

\email{cl114@rice.edu}

\author{Yingyan Lin}
\affiliation{%
  \institution{Rice University}
  \country{}
}
\email{yingyan.lin@rice.edu}

\begin{abstract}

The recent breakthroughs of deep neural networks (DNNs) and the advent of billions of Internet of Things (IoT) devices have excited an explosive demand for intelligent IoT devices equipped with domain-specific DNN accelerators. However, the deployment of DNN accelerator enabled intelligent functionality into real-world IoT devices still remains particularly challenging. First, powerful DNNs often come at prohibitive complexities, whereas IoT devices often suffer from stringent resource constraints. Second, while DNNs are vulnerable to adversarial attacks especially on IoT devices exposed to complex real-world environments, many IoT applications require strict security. Existing DNN accelerators mostly tackle only one of the two aforementioned challenges (i.e., efficiency or adversarial robustness) while neglecting or even sacrificing the other. To this end, we propose a \textit{2-in-1 Accelerator}, an integrated algorithm-accelerator co-design framework aiming at winning both the adversarial robustness and efficiency of DNN accelerators. Specifically, we first propose a Random Precision Switch (RPS) algorithm that can effectively defend DNNs against adversarial attacks by enabling random DNN quantization as an in-situ model switch during training and inference. Furthermore, we propose a new precision-scalable accelerator featuring (1) a new precision-scalable MAC unit architecture which spatially tiles the temporal MAC units to boost both the achievable efficiency and flexibility and (2) a systematically optimized dataflow that is searched by our generic accelerator optimizer. Extensive experiments and ablation studies validate that our \textit{2-in-1 Accelerator} can not only aggressively boost both the adversarial robustness and efficiency of DNN accelerators under various attacks, but also naturally support instantaneous robustness-efficiency trade-offs adapting to varied resources without the necessity of DNN retraining. We believe our \textit{2-in-1 Accelerator} has opened up an exciting perspective for robust and efficient accelerator design.

\end{abstract}

\ccsdesc[500]{Computer systems organization~Neural networks}

\keywords{neural networks, model robustness, precision-scalable accelerators}



\maketitle

\section{Introduction}

Deep neural networks' (DNNs) performance breakthroughs and the advent of billions of Internet of Things (IoT) devices have triggered an increasing demand for DNN-powered intelligent IoT devices. However, DNNs' deployments into real-world IoT devices still remain challenging. First, powerful DNNs' prohibitive complexity stands at odd with the stringent resource constraints of IoT devices \cite{drw,10.1145/3210240.3210337,9028245}. Second, while DNNs are vulnerable to adversarial attacks, many IoT applications require strict security under dynamic and complex real-world environments \cite{9096397}. Therefore, techniques boosting both DNNs' efficiency and robustness are highly desired. 

To tackle the first challenge, various domain-specific DNN accelerators \cite{chen2016eyeriss,jouppi2017datacenter,10.1109/ISCA45697.2020.00082,10.1109/ISCA45697.2020.00073,7780065,8050797} have been developed to customize the algorithm-to-hardware mapping methods (i.e., dataflows) and micro-architecture \cite{9053977} towards the workloads of DNNs to achieve orders-of-magnitude acceleration efficiency improvement over general computing platforms. In parallel, various techniques have been proposed to defend DNNs against adversarial attacks, showing promising performance to address the aforementioned robustness challenge. Among them, adversarial training~\cite{shafahi2019adversarial,madry2017towards,wong2019fast,tramer2017ensemble}, which augments the training set with adversarial samples generated on-the-fly during training, is currently the most effective method. Furthermore, recognizing that both efficiency and robustness are critical to many DNN-powered intelligent applications, pioneering efforts~\cite{wang2020dnnguard, rouhani2018deepfense, gan2020ptolemy} attempt to defend against adversarial attacks within DNN accelerators. Nevertheless, the art of robustness-aware DNN accelerators is still in its infancy, and existing defensive accelerators against adversarial attacks rely on additional detection networks/modules to detect/defend adversarial samples during inference, thus inevitably compromising their accelerator efficiency.

Considering that quantized DNNs are very promising as efficient DNN solutions and also highly desirable in many IoT applications \cite{fu2021cpt,fu2020fractrain}, we first ask an intriguing question: ``\textit{Is it possible to leverage quantization to boost DNNs' robustness}?", despite the fact that quantized DNNs have been shown to degrade the models' adversarial robustness unless being equipped with sophisticated regularization schemes \cite{song2020improving}. This is inspired by (1)~\cite{cohen2019certified, li2018certified, wu2020adversarial} showed that random permutations on the inputs can certifiably defend DNNs against adversarial attacks, and (2)~\cite{wu2020adversarial} found that weight perturbations are a good complement for input perturbations, because they can narrow the robust generalization gap as weights globally influence the losses of all examples. 
We thus hypothesize that quantization noise can be leveraged to provide similar effects as permutations to the weights/activations and thus enhance DNNs' robustness, motivating our random precision switch (RPS) algorithm that wins both efficiency and robustness of quantized DNNs. Furthermore, motivated by the bottlenecks of existing precision-scalable accelerators, we further develop a new  accelerator to enhance the acceleration efficiency of RPS equipped DNNs. Specifically, we make the following contributions:

\vspace{-0.5em}
\begin{itemize}

\item We propose an integrated algorithm-accelerator co-design framework dubbed \textit{2-in-1 Accelerator}, aiming at winning both the adversarial robustness and acceleration efficiency of DNN accelerators.

\item \textit{2-in-1 Accelerator}'s algorithm: We provide a new perspective regarding the role of quantization in DNNs' robustness, and propose a Random Precision Switch (RPS) algorithm that can effectively defend DNNs against adversarial attacks by enabling random DNN quantization as an in-situ model switch during training and inference. RPS equipped DNNs with fixed-point precisions even outperform their full-precision counterparts' robustness. 

\sloppy
\item \textit{2-in-1 Accelerator}'s architecture: We develop a new precision-scalable accelerator featuring (1) a novel multiply-accumulate (MAC) architecture which spatially tiles the temporal MAC units to boost both the achievable efficiency and precision-scalable flexibility and (2) a systematically optimized dataflow searched by our generic accelerator optimizer, surpassing existing precision-scalable accelerators. 

\item  We perform a thorough evaluation of \textit{2-in-1 Accelerator}  on six DNN models and four datasets under various adversarial attacks, and find that our \textit{2-in-1 Accelerator}~achieves up to 7.58$\times$ better energy efficiency, 4.59$\times$/36.5$\times$ higher throughput over precision-scalable/robustness-aware accelerators, and up to 24.48\% improvement in robust accuracy. We believe that our \textit{2-in-1 Accelerator} framework has not only demonstrated 
an appealing and effective real-world DNN solution, but also  
opened up an exciting perspective for winning both robustness and efficiency in DNN accelerators

\end{itemize}

\vspace{-0.5em}
\section{2-in-1 Accelerator: Algorithm}
\label{sec:alg}
In this section, we present our RPS algorithm that can simultaneously boost DNNs' robustness and efficiency and thus serve as the algorithmic enabler of our 2-in-1 accelerator.

\subsection{Preliminaries of adversarial robustness}
\label{sec:preliminaries}
\cite{goodfellow2014explaining} finds that DNNs are vulnerable to adversarial attacks, i.e., applying a small permutation $\delta$ within a norm ball ($\|\delta\| \leq \epsilon$) to the inputs can mislead DNNs' decisions. For example, the adversarial permutation $\delta$ under the $\ell_{\infty}$ attack~\cite{goodfellow2014explaining} is generated by maximizing the objective:

\setlength{\abovedisplayskip}{3pt}
\setlength{\belowdisplayskip}{3pt}
\vspace{-1em}
\begin{equation} \label{eq:general_attack} 
    \max_{\|\delta\|_{\infty} \leq \epsilon} \ell(f_{\theta}(x+\delta), y)
\end{equation} 
\noindent where $\ell$ is the loss function, $\theta$ is the weights of a DNN $f$, $x$ and $y$ are the input and the corresponding label, respectively.

To boost DNNs' robustness against adversarial attacks, adversarial training optimizing the following minimax problem is currently the strongest defense method~\cite{athalye2018obfuscated}:

\setlength{\abovedisplayskip}{3pt}
\setlength{\belowdisplayskip}{3pt}
\begin{equation} \label{eq:adversarial_training}
    \min_{\theta} \,\, \sum_{i} \max_{\|\delta\|_{\infty} \leq \epsilon} \ell(f_{\theta}(x_i+\delta), y_i)
\end{equation}
\vspace{-0.5em}






\vspace{-1em}
\subsection{Inspirations from previous works}
\label{sec:inspiration}

Previous works show that random smoothing or transformations~\cite{cohen2019certified, li2018certified, xie2017mitigating, guo2017countering} on the inputs help robustify DNNs and \cite{wu2020adversarial} shows that weight perturbations are good complements for input perturbations as they globally influence the learning loss of all inputs. Following this spirit, \cite{he2019parametric, dhillon2018stochastic, wu2020adversarial} explicitly introduce randomness and permutations in the models' weights or activations. On the other hand, \cite{strauss2017ensemble, tramer2017ensemble, liu2018towards} show that model ensemble can help improve robustness at a cost of efficiency due to the required multiple models. These two aspects inspire us to rethink the connection between quantization's role in the permutations of DNN weights/activations and model robustness and to view a DNN model under different precisions as an in-situ ensemble to boost both robustness and efficiency. As introduced in Sec.~\ref{sec:rps_alg}, the proposed RPS algorithm can be seen as an in-situ model switch among different precision choices.


\vspace{-0.5em}
\subsection{Poor transferability between precisions}
\label{sec:transferability}
To validate our above hypothesis that a DNN model under different precisions can be seen as an in-situ ensemble, we empirically check the robustness of such an ensemble by evaluating the transferability of adversarial attacks between different precisions. As elaborated below, we find that the adversarial attacks transfer poorly between different precisions of an adversarially trained model, regardless of its adversarial training methods and attack schemes.

\begin{figure}[!t]
\begin{center}
   \includegraphics[width=0.48\textwidth]{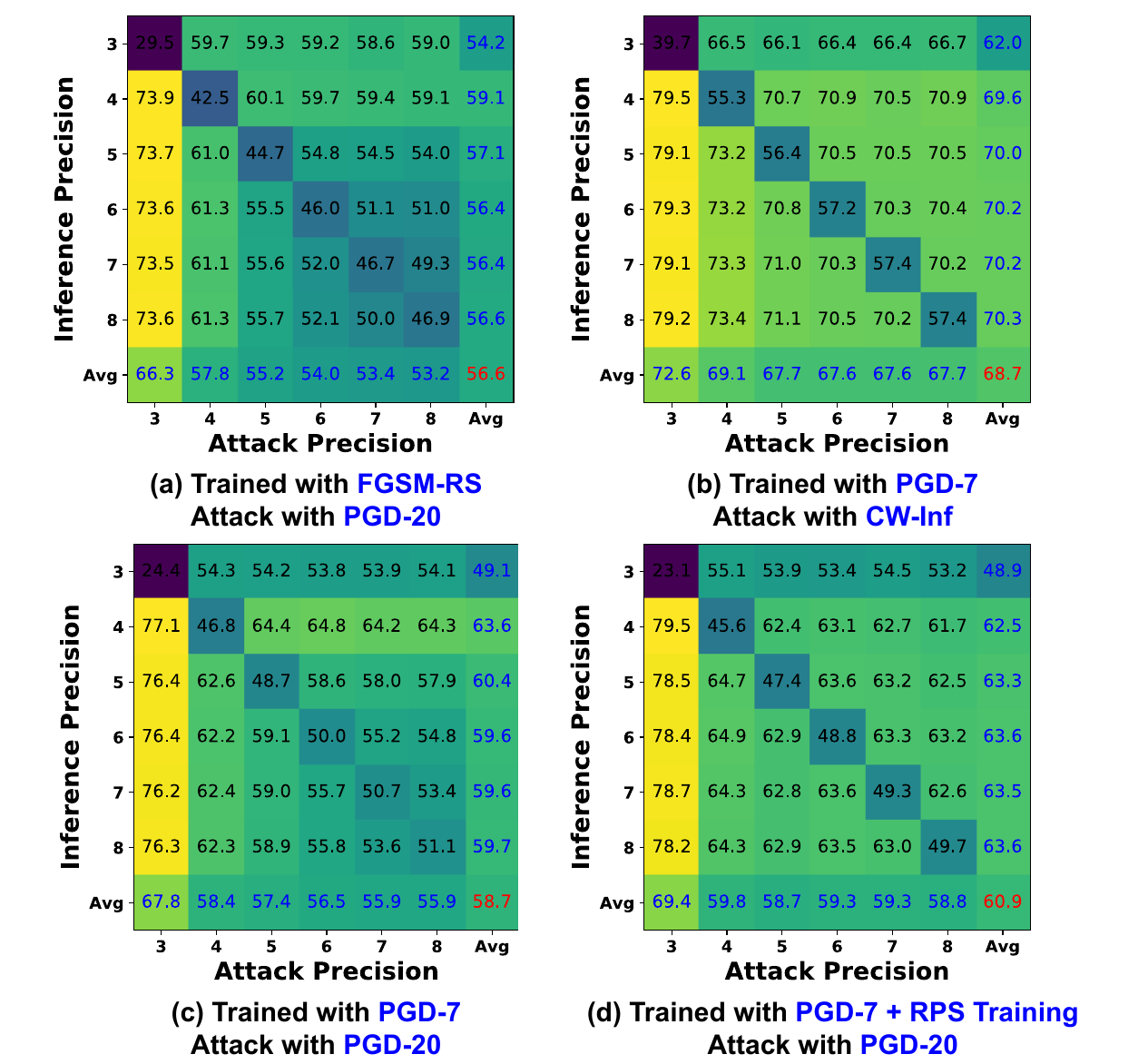}
\end{center}
  \vspace{-1.5em}
   \caption{Visualizing the transferability of adversarial attacks between different precisions, where the robust accuracy under different training methods (PGD-7 and FGSM-RS) and attacks (PGD-20 and CW-Inf) is annotated.}
   \vspace{-2em}
\label{fig:observation}
\end{figure}

\sloppy
\textbf{Experiment settings.} We conduct experiments that adopt adversarial attacks generated under one precision to attack the same adversarially trained model quantized to another precision. In particular,
we apply PGD-20~\cite{madry2017towards} and CW-Inf~\cite{carlini2017towards} attacks, to PreActResNet-18 (following~\cite{wong2019fast}) which is adversarially trained using different adversarial training methods~\cite{wong2019fast, madry2017towards} using an 8-bit linear quantizer~\cite{jacob2018quantization} under training settings introduced in Sec.~\ref{sec:exp_setup}. 
We annotate the robust accuracy evaluated on adversarial examples in Fig.~\ref{fig:observation} where the attack precision denotes the precision for generating attacks which are adopted to attack the same model quantized to another inference precision. The diagonal elements are the robust accuracy with the same attack/inference precision and the non-diagonal elements are the robust accuracy under transferred attacks from different precisions.

\textbf{Observations.} As observed from Fig.~\ref{fig:observation} (a)$\sim$(c), we can find that \underline{(1)} training and attacking at the same low precisions indeed notably degrade the robust accuracy, as shown in the diagonals of Fig.~\ref{fig:observation}, aligning with observations in~\cite{lin2019defensive}; \underline{(2)} it's more difficult for adversarial attacks generated under one precision to fool the same adversarially trained model quantized to a different precision, regardless of the relative difference between the two precisions; \underline{(3)} the poor transferability is consistent across different adversarial training and attack methods; and \underline{(4)} the average robust accuracies of all precisions under white-box attacks are consistently higher than the full-precision models trained with the corresponding adversarial training methods, indicating that randomly selecting an inference precision can potentially provide effective defense.The \textit{full-precision} accuracies of PreActResNet-18 trained with PGD-7/FGSM-RS are 51.2\%/47.1\%, respectively.

\textbf{Analysis.}
The key conclusion is that for white-box attacks, adversarial attacks generated at one precision transfer poorly to another precision. We hypothesize that this poor transferability is because adversarial perturbations are shielded by the quantization noise between the two precisions, which can not be effectively learned by gradient-based attacks.

\vspace{-0.5em}
\subsection{RPS towards robust DNNs}
\label{sec:rps_alg}
Motivated by the poor transferability between different precisions of a trained model, we propose the RPS algorithm to boost both model robustness and efficiency via enabling random DNN quantization as an in-situ model switch during training and inference.

\textbf{RPS training.} We propose the RPS training pipeline to (1) maintain a decent natural accuracy when the model is directly quantized to different precisions during inference, and (2) further increase the difficulty of transferring adversarial examples between different precisions. To this end, we adversarially train a model from scratch via (1) randomly selecting a precision from a candidate set in each iteration for generating adversarial examples and updating the model with the selected precision, and (2) equipping the model with switchable batch normalization (SBN)~\cite{jin2020adabits, guerra2020switchable} to independently record the statistics of different precisions given their corresponding adversarial examples. In particular, randomly selecting a training precision improves the capability of instant precision switch during inference and SBN enlarges the gap between different inference/attack precisions inspired by~\cite{jin2020adabits, guerra2020switchable, xie2020adversarial} which separately handles the statistics of different inputs. As shown in Fig.~\ref{fig:observation} (d), the same adversarially trained model equipped with RPS training shows larger robust gaps between different inference/attack precisions, especially under larger precision, as compared to the corresponding ones in Fig.~\ref{fig:observation} (c). Note that during inference, the multiplication and addition operations of SBN can be fused into the scale factors of linear quantizers~\cite{jacob2018quantization} and the model bias, respectively, thus does not require additional modules over existing low precision accelerators.

\textbf{RPS inference.} Given a model adversarially trained via our RPS training scheme, the proposed RPS inference randomly selects one precision from an inference precision set to quantize the model's weights and activations during inference. Based on the analysis in Sec.~\ref{sec:transferability}, randomly selecting an inference precision can greatly degrade the effectiveness of adversarial attacks as long as the attacks are not generated under the same precision, as consistently observed in Figs.~\ref{fig:observation}.

The RPS training and inference algorithms on top of PGD-7~\cite{madry2017towards} adversarial training are summarized in Alg.~\ref{alg:rps}, which is similar when applying on top of other adversarial training methods.

\vspace{-1em}
\subsection{Instant trade-offs between robustness and efficiency}
\label{sec:instant}

In addition to winning both robustness and efficiency, another benefit of our RPS algorithm is the instant trade-off capability between DNNs' robustness and efficiency during run-time to adapt to (1) the safety conditions of the external environments and (2) the remaining resource (e.g., battery power) on the device. Specifically,
our RPS achieves this via (1) switching to lower precisions when enabling random precision inference to trade robustness in less dangerous environments for a higher average efficiency, or (2) directly adopting a static low precision training under safe environments to pursue merely high efficiency.
This property can be highly desirable in real world applications especially intelligent IoT ones. We will next discuss the proposed accelerator that can not only improve the execution efficiency of DNNs resulting from our RPS algorithm but also set a new record of precision-scalable acceleration.

\setlength{\textfloatsep}{6pt}
\begin{algorithm}[t!]
\caption{The RPS Algorithm}
\begin{algorithmic}[1]
    \REQUIRE Training dataset $D_{train}$, model $f_{\theta}$, precision set $Set_{Q}$, total training epochs $T$, step size $\alpha$, adversarial dataset $D_{adv}$ generated on $f_{\theta}$ by attackers
    
    \STATE === RPS Training ===

    \STATE Equip $f_{\theta}$ with SBN
    
    \FOR{$epoch \in [1, T]$}
        \FOR{$(x,y) \in D_{train}$}
            \STATE Randomly select a precision $q$ from $Set_{Q}$
            \STATE Obtain $f^{q}_{\theta}$ by quantizing $f_{\theta}$ to $q$-bit

            \STATE $\delta = 0$ or random initialized
            \FOR{$t \in [1, 7]$}
                \STATE $\delta = clip_{\epsilon}\{\delta + \alpha \cdot sign(\nabla_{\delta} \ell(f^{q}_{\theta}(x+\delta), y))\}$
            \ENDFOR
        
            \STATE $\theta = \theta - \nabla_\theta \ell(f^{q}_\theta(x + \delta), y)$   
        \ENDFOR
    \ENDFOR
    
    \STATE === RPS Inference ===
    
    \FOR{$x_{adv} \in D_{adv}$}
        \STATE Randomly select a precision $q$ from $Set_{Q}$
        \STATE Obtain $f^{q}_{\theta}$ by quantizing $f_{\theta}$ to $q$-bit
        \STATE Evaluate $\hat{y} = f^{q}_{\theta}(x_{adv})$
    \ENDFOR
    \RETURN $\{\hat{y}\}$
\end{algorithmic}
\label{alg:rps}
\end{algorithm}

\section{2-in-1 Accelerator: Architecture}
\label{sec:accelerator}
In this section, we introduce our proposed accelerator architecture dedicated for variable-precision DNNs (e.g., RPS equipped DNNs in Sec.~\ref{sec:alg}) to achieve much improved acceleration efficiency. 
In particular, we first identify and analyze the bottlenecks of existing precision-scalable accelerators in Sec.~\ref{sec:bottleneck}, then present a new MAC unit architecture in Sec.~\ref{sec:mac_unit} and an automated accelerator optimizer in Sec.~\ref{sec:optimizer} that together tackles the aforementioned bottlenecks.



\vspace{-1em}
\subsection{Bottlenecks of SOTA precision-scalable accelerators}
\label{sec:bottleneck}

Despite the impressive performance achieved by SOTA precision-scalable accelerators~\cite{judd2016stripes, sharify2018loom, lee2018unpu, sharma2018bit, ryu2019bitblade, moons2017dvafs, moons20160, rzayev2017deeprecon}, they are still limited in their acceleration performance especially when accelerating more complex variable-precision DNNs, e.g., RPS equipped DNNs in which all the layers may switch their precision to any possible precision in a candidate set during inference. The bottlenecks of SOTA precision-scalable accelerators are described below. 

\vspace{-0.5em}
\subsubsection{Dilemma between flexibility and performance}
\label{sec:bottleneck_trade-off}

\textbf{\\ \indent Bottleneck.} While variable-precision DNNs have gained growing interest thanks to their advantages of enabling instantaneous energy-accuracy trade-off which is highly desirable in many DNN applications such as DNN-powered IoT ones, existing precision-scalable accelerators still struggle in the dilemma between their favored flexibility (i.e., support a large set of precisions) and achieved acceleration performance.

\textbf{Analysis.} SOTA precision-scalable accelerators can be categorized into two classes, i.e., \textit{temporal} and \textit{spatial} designs. The temporal designs~\cite{judd2016stripes, sharify2018loom} adopt bit-serial MAC units to execute a part of the bit operations between two operands during each cycle and then accumulate the results temporally via additional shift logic circuits to support variable precision inference; while the spatial architectures~\cite{sharma2018bit, moons2017dvafs} first split the execution of high precision multiplications into several 2-bit multipliers, and then exploit combinational logic circuits to dynamically compose and decompose the 2-bit multipliers to construct variable-precision MAC units. Both designs have their advantages and disadvantages:

On the one hand, temporal designs are inferior in their achieved throughput under lower precisions (<8-bit) compared with spatial designs as validated in~\cite{camus2019review}, since the area of their required shifters and accumulators are determined by their supported highest precision and thus can dominate the area cost, limiting their efficiency normalized over area~\cite{sharma2018bit}. 
On the other hand, spatial designs can only support a limited set of predefined precisions, e.g., 2-/4-/8-/16-bit for Bit Fusion~\cite{sharma2018bit}, if considering an affordable cost for their required configurability logic circuits due to the spatial constraints of their MAC units, while the precision choices in temporal designs are more flexible as higher precisions can naturally be supported by using more temporal cycles. Therefore, there exists a dilemma between the achieved flexibility and efficiency in SOTA precision-scalable accelerators.

\begin{figure}
    \centering
    \includegraphics[width=\linewidth]{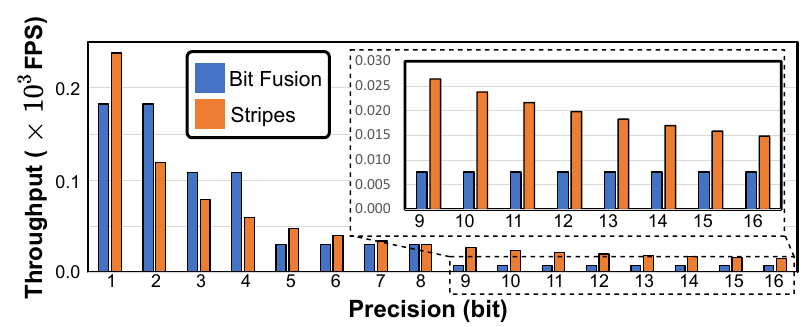}
    \vspace{-2em}
    \caption{Throughput under different precisions of Bit Fusion and Stripes for accelerating ResNet-50 on ImageNet.}
    \label{fig:method_throughput_bit}
\end{figure}

\textbf{Validation.} To validate the above analysis, we show the throughput under different precisions (the same for weights and inputs) of two representative spatial and temporal precision-scalable accelerators (i.e., Bit Fusion~\cite{sharma2018bit} and Stripes~\cite{judd2016stripes}) in Fig.~\ref{fig:method_throughput_bit}, when accelerating ResNet-50 on ImageNet. The detailed simulation settings can be found in Sec.~\ref{sec:exp_setup}. We can observe that (1) Bit Fusion achieves a higher throughput compared with Stripes under its \textit{supported} precisions (i.e., $<$8-bit, the most commonly adopted precisions in quantized DNNs~\cite{jung2019learning, bhalgat2020lsq+, esser2019learned, park2020profit}); (2) Bit Fusion leads to under-utilization of the hardware resources under its \textit{unsupported} precisions where it has to adopt the closest supported but higher precision; (3) Bit Fusion shows inferior throughput under precisions larger than 8-bit since it has to execute each Bit Bricks four times when the operands' precision is higher than 8-bit.
In contrast, while the temporal design, Stripes, is inferior to Bit Fusion under Bit Fusion's supported low precisions, it scales well with the precision, e.g., a consistent improvement in throughput  as the execution precision decreases.
This set of experiments demonstrates that SOTA precision-scalable accelerators inevitably suffer from the dilemma to trade-off between their achieved flexibility and efficiency, motivating our proposed new accelerator.

\vspace{-0.5em}
\subsubsection{Heavy shift-add overhead for variable-precision}
\label{sec:bottleneck_shift}

\textbf{\\ \indent Bottleneck.} To support variable-precision configurability, existing precision-scalable accelerators require a heavy shift-add overhead, e.g., the shifters in the bit-serial units of the temporal designs~\cite{judd2016stripes} and the shifters for composing various 2-bit multipliers in the spatial designs~\cite{sharma2018bit} introduce significant or even dominant area and energy costs.

\textbf{Analysis from related works.} The size of the required shifters and accumulators in the \textit{temporal designs} are determined by its highest supported precision and thus can dominate the area cost~\cite{sharma2018bit}, e.g., the shifter and the accumulator use up around 90\% of the total area in a temporal design supporting up to 16-bit, greatly limiting their achievable benefits and leading to inferior normalized efficiency per area. Similar observations have been drawn in~\cite{camus2019review} that compared with spatial designs, temporal designs have a worse normalized performance, i.e., throughput/area. On the other hand, for \textit{spatial designs}, \cite{camus2019review} shows that their MAC unit can require up to 4.4× the area of a standard MAC unit due to the overhead of their scalable units using sub-computation parallelism, and \cite{ryu2019bitblade} also finds that the shift-add logic circuit in Bit Fusion for supporting precision-scalable configuration occupies a surprisingly large
area (67\%) and consumes a majority of power consumption (79\%).
These observations motivate us to explore a new precision-salable accelerator to reduce the shift-add logic overhead and thus to better allocate the limited area for more MAC units. 

\begin{figure}
    \centering
    \includegraphics[width=0.8\linewidth]{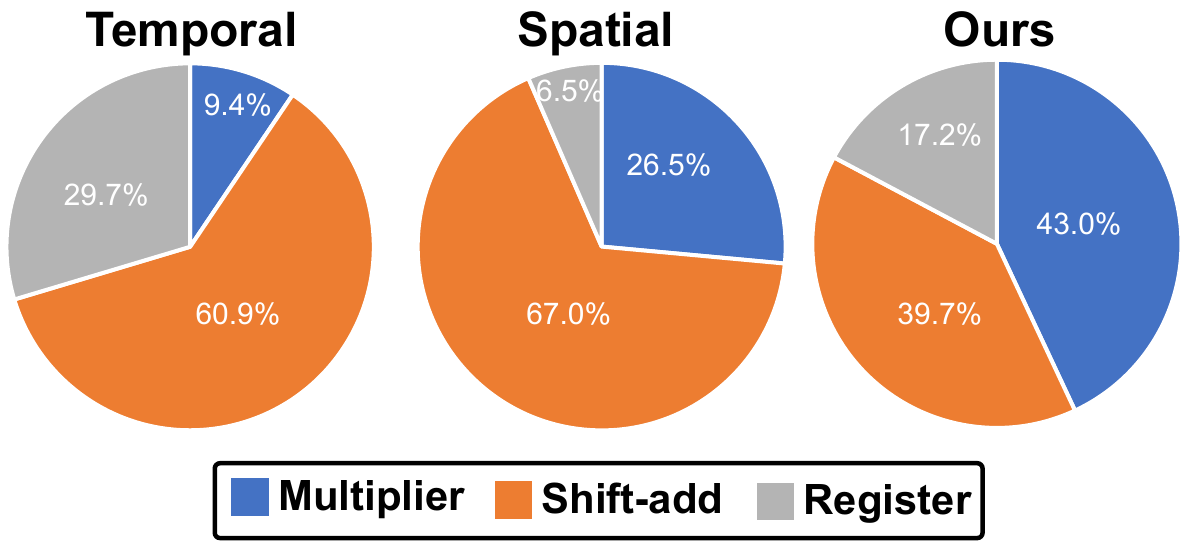}
    \vspace{-0.5em}
    \caption{Area breakdown of the MAC units based on SOTA temporal/spatial designs and our proposed design.}
    \label{fig:area_breakdown}
\end{figure}

\textbf{Validation.} 
In Fig.~\ref{fig:area_breakdown}, we show the area breakdown of the MAC units in Bit Fusion~\cite{sharma2018bit}, a temporal design reported by Bit Fusion, as well as our proposed architecture introduced in Sec.~\ref{sec:accelerator}. We can see that the shift-add logic occupies 60.9\%/67.0\% of the total area in the MAC units of the temporal/spatial designs. In contrast, our design reduces the area of shift-add logic to 39.7\% via the techniques proposed in Sec.~\ref{sec:accelerator}, thus leading to a better performance/area.

\vspace{-0.5em}
\subsubsection{Fixed or limited dataflow optimization}
\label{sec:bottleneck_mapping}

\textbf{\\ \indent Bottleneck.} The dataflow of DNN accelerators largely impacts their acceleration efficiency \cite{meng2017magnet,10.1145/3373087.3375306,9502478,fu2021autonba}. For variable-precision DNNs (e.g., RPS equipped ones), each layer might be executed at any precision of the candidate precision set, making it more challenging to find an optimal dataflow for all the cases. For example, a 20-layer DNN with 5 precision choices correspond to a total of 100 different dataflows to be explored for achieving the best average efficiency, in contrast to only 20 for its static layer-wise mixed-precision counterpart. 

\textbf{Analysis.} 
As analyzed in Eyeriss~\cite{chen2016eyeriss, zhang2020dna,9502478,fu2021instantnet}, dataflows can be described as the tiling strategies, including the loop order and tiling factors, across each memory hierarchy. Most of existing precision-scalable accelerators adopt a fixed dataflow within their memory hierarchies or only conduct a limited dataflow optimization. In particular, \cite{judd2016stripes, sharify2018loom, ryu2019bitblade} all use a fixed NoC (Network-on-Chip as defined in~\cite{chen2016eyeriss}) design, i.e., fixing the tiling strategies along both the two dimensions of the MAC array; and Bit Fusion~\cite{sharma2018bit} provides a dataflow optimization tool which only considers the loop order optimization for the global buffer and thus lacks flexible dataflow support for other memory hierarchies. Considering that different networks/layers with different precisions might favor different dataflows, a more comprehensive optimizer is necessary to find the optimal dataflow for further boosting the efficiency of precision-scalable accelerators.

\vspace{-0.5em}
\subsection{The proposed MAC unit architecture}
\label{sec:mac_unit}
In this subsection, we introduce the proposed MAC unit architecture. Specifically, we show how a vanilla spatial-temporal MAC unit architecture in Sec.~\ref{sec:spatial-temporal} is evolved into our proposed MAC unit architecture step by step through the optimization methods in Sec.~\ref{sec:reorganize} $\sim$~\ref{sec:fuse-bit-serial}, and finally present the overall accelerator architecture in Sec.~\ref{sec:final_arch}.

\subsubsection{A spatial-temporal design}
\label{sec:spatial-temporal}

\begin{figure}
    \centering
    \includegraphics[width=\linewidth]{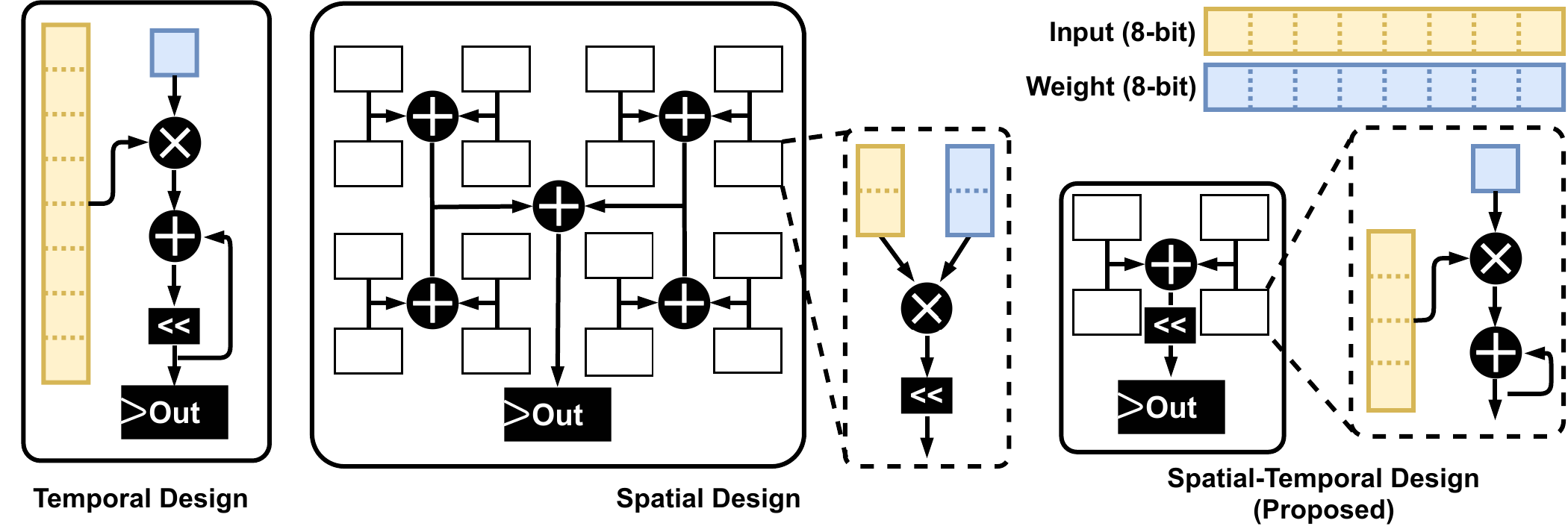}
    \vspace{-2em}
        \caption{The MAC unit of the temporal design, spatial design, and our spatial-temporal design which spatially tiles the temporal units to marry the advantages of both temporal and spatial designs for variable precision execution. For 8-bit weight and input in this case, it takes 8 cycles, 1 cycles, and 4 cycles for the temporal, spatial, and our design.}
    \label{fig:spatial_vs_temp}
\end{figure}

\textbf{\\ \indent Key idea.} As analyzed in Sec.~\ref{sec:bottleneck_trade-off}, there exists an inevitable trade-off between bit-level flexibility and acceleration efficiency in temporal and spatial designs. As both flexibility and efficiency are critical for real applications, we propose a new spatial-temporal MAC unit architecture which spatially tiles the temporal units to combine the advantages of both temporal and spatial designs. In Fig.~\ref{fig:spatial_vs_temp}, we show an overview of the MAC unit in the (a) temporal, (b) spatial, and (c) our proposed spatial-temporal designs. We tile the temporal units, i.e., bit-serial units, in the same manner as the Bit Bricks in Bit Fusion~\cite{sharma2018bit} so that they can be dynamically composed to support variable precisions, e.g., each of the four bit-serial units takes four cycles to calculate a 2-bit $\times$ 2-bit partial product, the results of which are then fused via shift and accumulation to obtain the final 4-bit $\times$ 4-bit results.


\textbf{Advantages of the spatial-temporal design.} \underline{First}, our spatial-temporal design maintains a high flexibility in the execution precision choices. Spatial designs~\cite{sharma2018bit} can only support limited precision choices (like 2-/4-/8-/16-bit) while our design can flexibly support more commonly used precision, e.g, each of the four bit-serial units can take three cycles to calculate a total of four 3-bit $\times$ 3-bit products, or one 6-bit $\times$ 6-bit product via dynamic composition.
\underline{Second}, the smaller size (i.e., the supported maximal precision) of the bit-serial units in our spatial-temporal design will help mitigate the area bottleneck caused by the shift-add logic for precision configuration. In particular, one major bottleneck of temporal designs when supporting a high bit-level flexibility is that their shifters and accumulators within each bit-serial module are determined by their highest supported precision, e.g., dominating a 90\% of the area in a 16-bit bit-serial unit, as pointed out by~\cite{sharma2018bit}. Our spatial-temporal design tackles this bottleneck via spatially composing bit-serial units of smaller sizes, i.e., each bit-serial unit can support up to 4-bit $\times$ 4-bit to constrain the maximal size required by the shifters. \underline{More importantly}, the number of the required shift-add logic within the bit-serial unit and between different units for dynamic composition can be aggressively reduced with further optimization as introduced in Sec.~\ref{sec:reorganize} $\sim$~\ref{sec:fuse-bit-serial}.

Note that Bit Fusion also adopts a temporal-spatial manner for 16-bit inference by temporally executing 8-bit inference with their spatial unit for four cycles to compose a 16-bit result to avoid more complex logic for precision configurability, e.g., shifters of larger sizes. However, their temporal execution of the spatial units cannot benefit the bit-level flexibility like our design which spatially tiles the temporal units.

\textbf{Spatial-temporal scheduling for different precisions.}
In our design, each bit-serial unit supports up to 4-bit $\times$ 4-bit calculation and each MAC unit adopts up to four bit-serial units for calculating one partial sum, i.e., supporting up to 8-bit $\times$ 8-bit calculation. For dealing with the precision higher than 8-bit, we follow Bit Fusion to temporally execute the whole MAC unit and then accumulate their results, considering that (1) the cost of more complex precision configurability under higher precisions will be higher and (2) 8-bit or lower precisions are sufficient for most DNN inference without accuracy degradation~\cite{jung2019learning, bhalgat2020lsq+, esser2019learned, park2020profit}.


Next, we introduce the detailed schedule of our MAC unit that is conducted spatially across the bit-serial units and temporally across cycles under each precision. Specifically, for operands with precisions no more than 4-bit, each bit-serial unit will independently calculate one partial sum of the final output; For operands with up to 6-bit $\times$ 6-bit / 8-bit $\times$ 8-bit, each of the four bit-serial units calculates a partial product with up to 3-bit $\times$ 3-bit / 4-bit $\times$ 4-bit, and then all the partial products will be composed to the final result via shift and accumulation; For operands with more irregular precisions like 5-bit $\times$ 5-bit, we split it into (3-bit+2-bit)$\times$(3-bit+2-bit), i.e., four bit-serial units will take the computation of 3-bit $\times$ 3-bit, 2-bit $\times$ 2-bit, and two 3-bit $\times$ 2-bit, respectively, and similarly, operands with 7-bit can be split into (4-bit+3-bit); and for operands higher than 8-bit, the calculation will be split to no more than 8-bit and temporally executed by the whole MAC unit as mentioned above, e.g., 12-bit $\times$ 12-bit can be split into four 6-bit $\times$ 6-bit, each of which will be sequentially executed by the MAC unit and then accumulated. The above analysis also works for asymmetrical precisions, e.g., 4-bit $\times$ 2-bit which takes only two cycles for each bit-serial units to complete the execution.

\vspace{-0.5em}
\subsubsection{Opt-1: Reorganize bit-level split/allocation}
\label{sec:reorganize}


\begin{figure}
    \centering
    \includegraphics[width=\linewidth]{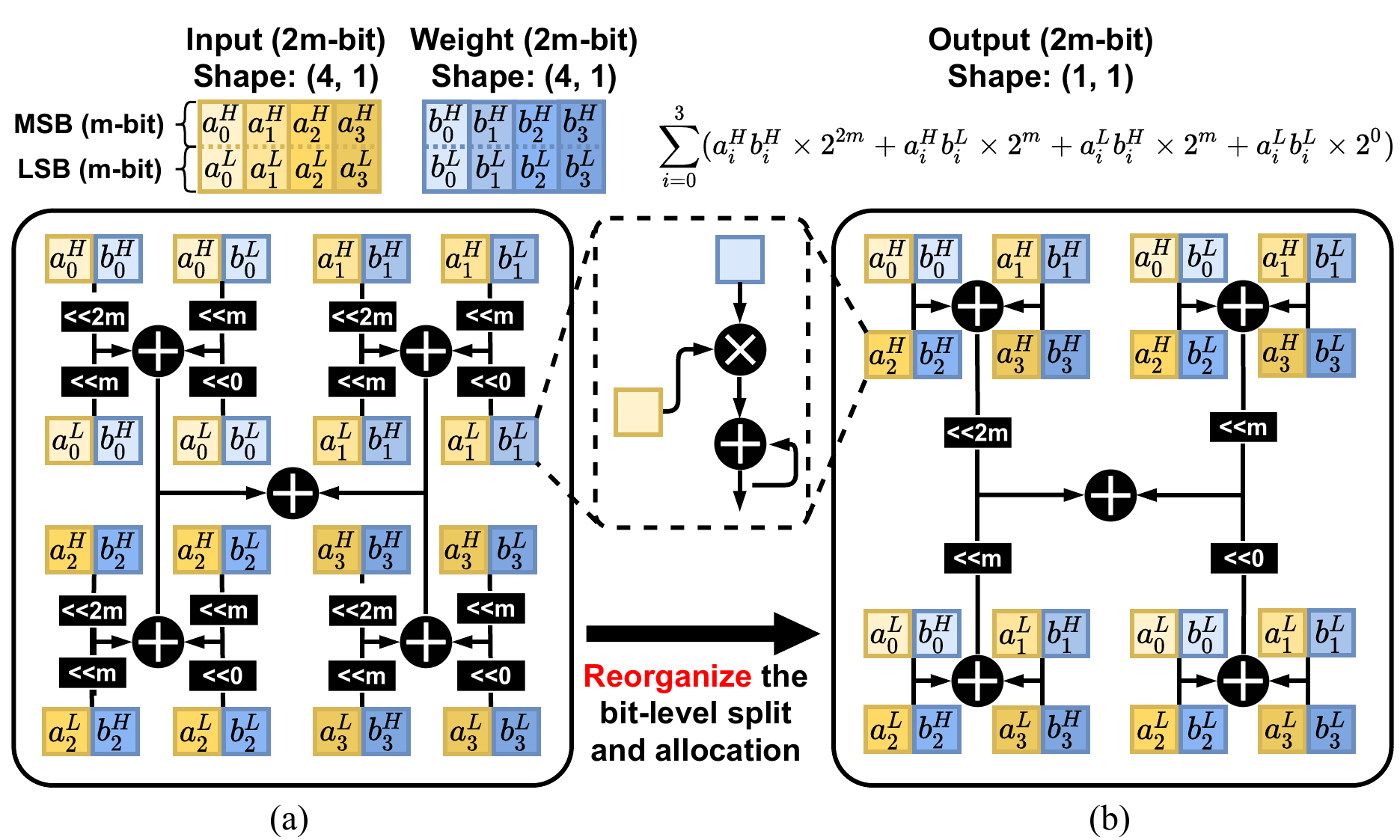}
    \vspace{-2em}
    \caption{Reorganizing the bit-level split and allocation reduces the number of shifters by 1/$n$ ($n$=4 in this case, denoting the number of partial sums) when handling the inputs and weights of 2m-bit. Here $a_{i}^{L}$/$b_{i}^{L}$ is the first m-bit LSB of inputs/weights and $a_{i}^{H}$/$b_{i}^{H}$ is the remaining MSBs of the $i$-th partial sum.}
    \vspace{-0.5em}
    \label{fig:reorganize}
\end{figure}

\textbf{\\ \indent Motivation.} It's important to improve bit-level split and allocation of the inputs/weights for the MAC units in precision-scalable accelerators, considering that the overhead of the shifters and accumulators for precision configurability is coupled with the workload patterns~\cite{camus2019review}. For example, if each bit-serial unit in a MAC unit processes one bit-level partial product of different outputs, their outputs need to be accumulated by different accumulators, thus requiring a large area overhead. Therefore, we aim to reorganize the workloads, more specifically, the bit-level split and allocation strategy to reduce the required shifters and accumulators in a MAC unit.

\textbf{Calculating multiple partial sums in one MAC unit.} We increase the number of bit-serial units in each MAC unit to simultaneously calculate multiple partial sums of the same output pixel as shown in Fig.~\ref{fig:reorganize} (a), which implies that the weights come from different kernel rows (R) and columns (S) while the inputs come from different input channels (C) for calculating the partial sums. Therefore, all the partial sums can be directly accumulated in one accumulator regardless of the execution precision. 
From a tiling strategy perspective, we explicitly tile the R, S, or C dimension in the MAC unit for further improving the area/energy efficiency while freeing up the used dataflow in the NoC (i.e., MAC array) and global buffer levels for layerwise optimization as introduced in Sec.~\ref{sec:optimizer}. Such a flexibility is necessary for dataflow optimization towards reducing the data movement cost of each layer. \underline{More importantly}, simultaneously calculating multiple partial sums also brings out another opportunity to aggressively reduce the required shifters as introduced below.

\textbf{Reorganize the bit-level split and allocation.} The number of shifters for the dynamic composition of bit-serial units can be reduced via reorganizing the bit-level split and allocation strategy. Suppose that calculating the $i$-th partial sum of an operand $a_i$ can be formulated as $a_i = a_i^H \times 2^m + a_i^L \times 2^0$ where $a_i^L$ is the first $m$-bit LSB and $a_i^H$ is the remaining MSBs, then the final result of one MAC unit can be formulated as the sum of the totally $n$ partial sums:

\setlength\abovedisplayskip{0pt}
\setlength\belowdisplayskip{0pt}
\setlength{\abovedisplayshortskip}{0pt}
\setlength{\belowdisplayshortskip}{0pt}
\begingroup
\tiny{
\begin{align}
    \begin{split}
    \sum^{n-1}_{i=0} &(a_i^H \times 2^m + a_i^L \times 2^0)(b_i^H \times 2^m + b_i^L \times 2^0)
    \label{eq:vanilla} \end{split}\\
    \begin{split}
    = \sum^{n-1}_{i=0} &(a_i^H b_i^H \times 2^{2m} + a_i^H b_i^L \times 2^m + a_i^L b_i^H \times 2^m + a_i^L b_i^L \times 2^0) \label{eq:original}
    \end{split}\\
    \begin{split}
    = \sum^{n-1}_{i=0} &(a_i^H b_i^H) \times 2^{2m} + \sum^{n-1}_{i=0}(a_i^H b_i^L) \times 2^m + \sum^{n-1}_{i=0} (a_i^L b_i^H) \times 2^m + \sum^{n-1}_{i=0} (a_i^L b_i^L) \times 2^0 \label{eq:reorganize}
    \end{split}
\end{align}
}
\endgroup

The original design in Fig.~\ref{fig:reorganize} (a) corresponds to Eq.~\ref{eq:original} where totally 4$n$ shifters are required for combining the outputs from different temporal units, whereas the reformulation in Eq.~\ref{eq:reorganize} only requires 4 shifters.
Inspired by this, instead of accumulating different partial sums, we adopt a \textit{first-reduce-then-shift} strategy that the partial products of the same magnitude (i.e., requiring the same number of shifts) from different partial sums are organized as a group which is mapped into a set of bit-serial units as shown in Fig.~\ref{fig:reorganize} (b). In this way, the outputs of the bit-serial units in a group can be directly summed together without any shifters and the final result of one MAC unit is the combination of the outputs from different groups via a group-wise shift-add logic. This is equivalent to accumulate different partial sums in Fig.~\ref{fig:reorganize} (a) as formulated in Eq.~\ref{eq:original} but the number of shifters \textbf{cross the bit-serial units} for precision configurability is reduced by 1/$n$ as shown in Eq.~\ref{eq:reorganize}.

\vspace{-0.5em}
\subsubsection{Opt-2: Fuse the shift-add logic of bit-serial units in one group}
\label{sec:fuse-bit-serial}

\begin{figure*}
    \centering
    \includegraphics[width=0.88\linewidth]{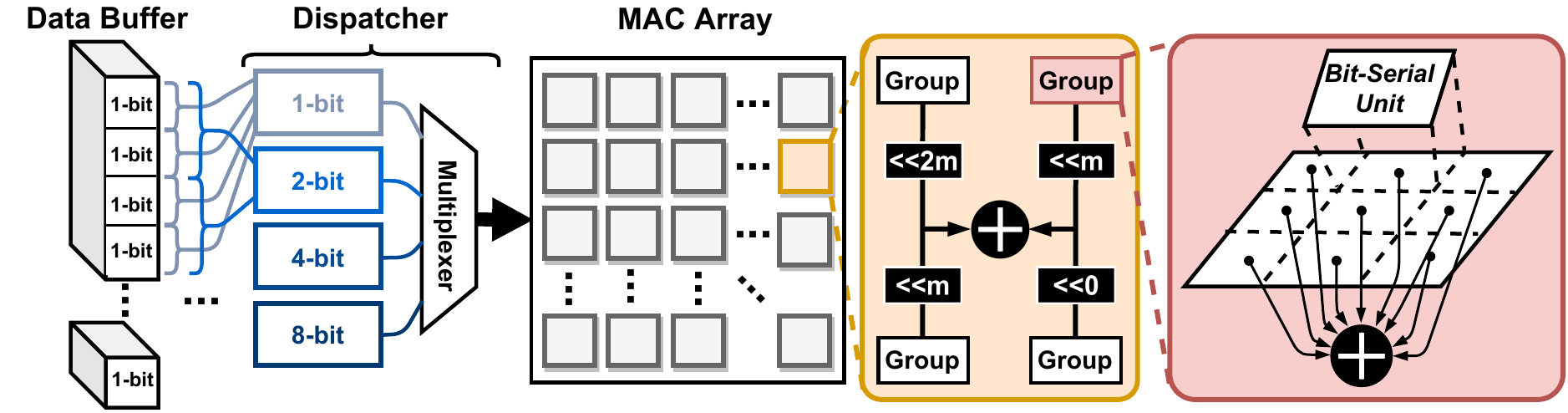}
    \vspace{-1em}
    \caption{The overall architecture of the proposed \textit{2-in-1 Accelerator}.}
    \vspace{-1.3em}
    \label{fig:overall_design}
\end{figure*}

As introduced in Sec.~\ref{sec:reorganize}, the outputs from each group of the bit-serial units can be directly accumulated without any shifter between the bit-serial units. This property brings another significant benefit in that all the shift-add logic of the bit-serial units in one group can be fused into one shift-add logic, named group shift-add, as shown in Fig.~\ref{fig:overall_design} (the leftmost zoom-in of one group). In particular, since the total number of shifts is the same for all the bit-serial units in one group, in each cycle the partial products of all the bit-serial units in one group can be directly summed together and then fed into the group shift-add module. Such an optimization reduces the required number of shifters \textbf{within the bit-serial units} by 1/$n$. 
The synthesized results show that our final MAC unit design in Fig.~\ref{fig:overall_design} achieves 2.3$\times$ and 4.88$\times$ improvement in throughput/area and energy-efficiency/operation, respectively, compared with Bit Fusion under 8-bit$\times$8-bit.

Note that (1) this optimization is specific to our design that organizes the bit-serial units into groups without the necessity of having unit-wise shifters and such opportunities do not exist in previous temporal/spatial designs; and (2) although the group shift-add can be potentially further combined with the group-wise shift-add, this will also increase the critical path and limit the system frequency. Thus, we keep them as two separate parts in our design.

\vspace{-0.5em}
\subsubsection{Overall architecture}
\label{sec:final_arch}


\textbf{\\} The overall \textit{2-in-1 Accelerator} architecture is shown in Fig.~\ref{fig:overall_design}, where the data is packed by a dispatcher which is implemented by a multiplexer to enable different granularities (i.e., 1/2/4/8-bit) for accessing the data buffer, and then passed to the MAC array as described in Sec.~\ref{sec:spatial-temporal}$\sim$~\ref{sec:fuse-bit-serial} for further processing. To this end, our \textit{2-in-1 Accelerator} can (1) fully achieve the ``win-win'' in robustness and efficiency on top of the proposed RPS algorithm, and (2) support instantaneous robustness-efficiency trade-offs as validated in Sec.~\ref{sec:exp_instant_trade_off}.

\vspace{-0.7em}
\subsection{The proposed automated optimizer}
\label{sec:optimizer}

It is well-known that both the dataflow and micro-architecture of a DNN accelerator are critical to its achievable efficiency. For example, \cite{meng2017magnet} shows that different dataflows can result in a 10× difference in the accelerators' efficiency. Meanwhile, the number of all possible dataflows and micro-architectures for an accelerator can easily explode \cite{autodnnchip}, which can be time-consuming and might not even be practical to manually identify, thus it can be greatly useful to have a generic accelerator optimizer that can automatically search for both the optimal dataflow and micro-architecture given the target acceleration efficiency and hardware resource (e.g., area).
To this end, we propose an automated optimizer with two modes, i.e., (1) search for merely dataflows and (2) search for both the dataflows and micro-architectures given an area budget.

\textbf{Searching for merely dataflows.} For this mode, we adopt an evolutionary algorithm~\cite{miller1995genetic}. Specifically, the searchable factors include the tiling factors for each data dimension and the loop order for each memory hierarchy. Note that the optimal refresh location, which is the one occupying the most memory size without causing overflow, can be automatically derived since all the memory sizes are fixed in this mode. If all possible refresh locations cause overflows, the corresponding design is invalid and discarded. As shown in Alg.~\ref{alg:dataflow}, we start from a population of randomly initialized for-loop descriptions and in each cycle, select the top 30\% designs in terms of efficiency as a new population, and then conduct (1) \underline{crossover} (i.e., generate a new design via randomly selecting two designs from the population and inserting one design's loop order in one memory hierarchy or tiling factors of one data dimension to the other design) and (2) \underline{mutation} (i.e., generate a new design from a randomly selected dataflow via randomly permuting its loop order in one memory hierarchy or tiling factors of one data dimension to another choice). After enlarging the pool to the original population size, we will start a new cycle and iterate this process until reaching a predefined maximal cycle number. Note that in both modes, we adopt an open-sourced generic performance predictor of DNN accelerators \cite{zhao2020dnnchip} to obtain the efficiency for a given dataflow and micro-architecture pair.

\textbf{Searching for both dataflows and micro-architectures.} The search engine under this mode can be built on top of that for the above mode. Specifically, we predefine a design space with a set of available choices for the MAC array size and memory sizes in each memory hierarchy which are then synthesized to acquire the unit energy and area; and then adopt another evolutionary algorithm similar to Alg.~\ref{alg:dataflow} to explore the design space, where the efficiency of an micro-architecture is measured by calculating its average energy/throughput under different  precisions after optimizing the dataflow via Alg.~\ref{alg:dataflow}.

\textbf{Note that} for a fair comparison with the baselines, in this work we only optimize the dataflow of each workload and adopt the same memory/MAC array area as our baselines.

\begin{algorithm}[t!]
\caption{Evolutionary Search for Dataflows}
\begin{algorithmic}[1]
    \REQUIRE Architecture $arch$, Workload (layer information and execution precision), Total cycle number $Total\_Cycle$, Population size $Psize$
    
    \STATE Initialize a $population$ of $dataflow$ with different loop orders and tiling factors according to the workload

    \FOR {$cycle \in [1, Total\_Cycle]$}
        \STATE Select the top 30\% $dataflow$ from the $population$ based on the predicted efficiency of the workload
        \WHILE {$size(population) < Psize$}
            \STATE Randomly select two $dataflow$, do \underline{crossover}, append to $population$ if valid
            \STATE Randomly select one $dataflow$, do \underline{mutation}, append to $population$ if valid
        \ENDWHILE
    \ENDFOR
    \RETURN The best $dataflow$ in the $population$
\end{algorithmic}
\label{alg:dataflow}
\end{algorithm}
\section{Experiment Results}
\subsection{Experiment Setup}
\label{sec:exp_setup}

\subsubsection{Algorithm Setup}
\textbf{\\ \indent Networks \& datasets.} We evaluate our RPS algorithm on \underline{three} networks and \underline{four} datasets which are the most commonly used ones in the robustness literature~\cite{madry2017towards, shafahi2019adversarial, wong2019fast}, i.e., PreActResNet-18 and WideResNet-32 on CIFAR-10/CIFAR-100/SVHN and ResNet-50 on ImageNet. We use a linear quantizer~\cite{jacob2018quantization} for quantizing weights/activations to the same precision.

\textbf{Training settings.} We adopt four SOTA adversarial training methods, including FGSM~\cite{goodfellow2014explaining}, FGSM-RS~\cite{wong2019fast}, PGD-7~\cite{madry2017towards}, and Free~\cite{shafahi2019adversarial} and apply our RPS algorithm on top of them. We follow their original papers for the adversarial training hyper-parameter settings and follow the model training settings in~\cite{madry2017towards} and~\cite{shafahi2019adversarial} for CIFAR-10/CIFAR-100/SVHN and ImageNet.

\textbf{Attack settings.} We consider the strong attacks including three white-box attacks PGD~\cite{madry2017towards}, AutoAttack~\cite{croce2020reliable}, CW~\cite{carlini2017towards}, and one gradient-free attack Bandits~\cite{ilyas2018prior}, with different numbers of iterations/restarts and permutation strengths $\epsilon=8, 12, 16$.
Without loss of generality, we assume the adversary adopts random precision from the same inference precision set as our RPS  since (1) any attack precision out of RPS's inference precision set will merely increase RPS's robust accuracy according to Fig.~\ref{fig:observation}, and (2) while the adversary may select precisions with better attacking success rates, our RPS can also increase the probability of sampling more robust precisions for a stronger defense, thus we assume both the adversary and RPS adopt random precision for simplicity.

\vspace{-0.5em}
\subsubsection{Accelerator Setup}
\textbf{\\ \indent Accelerator development and synthesis.}
In order to evaluate our proposed accelerator, we implement a custom cycle-accurate simulator, aiming to model the behavior of the synthesized circuits. The design parameters in the simulator are obtained from gate-level netlists and SRAM which are generated based on a commercial 28nm technology using the Synopsys Design Compiler and Memory compiler provided by the foundry. Specifically, proper activity factors are set at the input ports of the memory/computation units, and the energy is calculated using PrimeTime~\cite{PTPX}.

\textbf{Baselines.} We benchmark with two SOTA precision-scalable accelerators Bit Fusion~\cite{sharma2018bit} and Stripes~\cite{judd2016stripes}, and one robustness-aware accelerator DNNGuard~\cite{wang2020dnnguard}. For a fair comparison, we adopt the same memory area and MAC array area with Bit Fusion, and we modify the unit energy of Bit Fusion's official simulator to scale it from 45nm to 28nm following the rule in~\cite{scaling}. For Stripes, thanks to the clear description of the design in their paper and the easy representation, we built a cycle-accurate simulator for it with the same memory/MAC array area with Bit Fusion and our design, and optimize its dataflow with our automated optimizer.

\textbf{Workloads.} We adopt six networks (WideResNet-32/ResNet-18 on CIFAR-10 with 32$\times$32 inputs and AlexNet/VGG-16/ResNet-18/50 on ImageNet with 224$\times$224 inputs) under 1$\sim$16-bit as our workloads.

\subsection{Evaluate 2-in-1 Accelerator’s algorithm}
\label{label:exp_rps}
We evaluate the improvement in robustness via applying the proposed RPS on top of SOTA adversarial training methods. Note that all the baselines are SOTA adversarial training methods with a full precision, i.e., no quantization is applied. 
Our RPS adopts a precision set of 4$\sim$16-bit by default.

\begin{table}[t]
\centering
\caption{Evaluating RPS on two networks and three adversarial training methods FGSM~\cite{goodfellow2014explaining}, FGSM-RS~\cite{wong2019fast}, and PGD-7~\cite{madry2017towards} on CIFAR-10 under different PGD attacks.}
\vspace{-1em}
\resizebox{0.48\textwidth}{!}{
\begin{tabular}{ccccccc}
\toprule
\textbf{} & \multicolumn{3}{c}{\textbf{PreActResNet-18}} & \multicolumn{3}{c}{\textbf{WideResNet-32}} \\ \hline
\textbf{\begin{tabular}[c]{@{}c@{}}Adversarial \\ Training Method\end{tabular}} & \textbf{\begin{tabular}[c]{@{}c@{}}Natural \\ (\%)\end{tabular}} & \textbf{\begin{tabular}[c]{@{}c@{}}PGD-20\\  (\%)\end{tabular}} & \textbf{\begin{tabular}[c]{@{}c@{}}PGD-100 \\ (\%)\end{tabular}} & \textbf{\begin{tabular}[c]{@{}c@{}}Natural \\ (\%)\end{tabular}} & \textbf{\begin{tabular}[c]{@{}c@{}}PGD-20\\  (\%)\end{tabular}} & \textbf{\begin{tabular}[c]{@{}c@{}}PGD-100 \\ (\%)\end{tabular}} \\ \hline \hline
FGSM & 67.04 & 41.48 & 41.37 & 66.76 & 40.78 & 40.55 \\
FGSM \textbf{+ RPS} & 80.58 & \textbf{64.08} & \textbf{63.56} & 64.09 & \textbf{50.70} & \textbf{48.72} \\ \hline
FGSM-RS & 86.08 & 41.76 & 41.13 & 89.95 & 45.33 & 44.77 \\
FGSM-RS \textbf{+ RPS} & 82.11 & \textbf{59.33} & \textbf{59.32} & 87.87 & \textbf{60.07} & \textbf{59.12} \\ \hline
PGD-7 & 82.02 & 51.17 & 50.93 & 85.25 & 54.61 & 54.36 \\
PGD-7 \textbf{+ RPS} & 82.16 & \textbf{65.15} & \textbf{64.88} & 81.52 & \textbf{66.75} & \textbf{66.28} \\ \hline
\end{tabular}
}
\label{tab:sota_cifar10}
\vspace{-1em}
\end{table}

\begin{table}[t!]
\centering
\caption{Evaluating RPS on two networks trained with FGSM-RS~\cite{wong2019fast} and PGD-7~\cite{madry2017towards} on CIFAR-100.}
\vspace{-1em}
\resizebox{0.48\textwidth}{!}{
\begin{tabular}{ccccccc}
\toprule
\multicolumn{1}{l}{} & \multicolumn{3}{c}{\textbf{PreActResNet-18}} & \multicolumn{3}{c}{\textbf{WideResNet-32}} \\ \hline 
\textbf{\begin{tabular}[c]{@{}c@{}}Adversarial \\ Training Method\end{tabular}} & \textbf{\begin{tabular}[c]{@{}c@{}}Natural \\ (\%)\end{tabular}} & \textbf{\begin{tabular}[c]{@{}c@{}}PGD-20 \\ (\%)\end{tabular}} & \textbf{\begin{tabular}[c]{@{}c@{}}PGD-100 \\ (\%)\end{tabular}} & \textbf{\begin{tabular}[c]{@{}c@{}}Natural \\ Acc (\%)\end{tabular}} & \textbf{\begin{tabular}[c]{@{}c@{}}PGD-20 \\ (\%)\end{tabular}} & \textbf{\begin{tabular}[c]{@{}c@{}}PGD-100 \\ (\%)\end{tabular}} \\ \hline \hline
FGSM-RS & 57.6 & 26.14 & 25.88 & 67.29 & 25.35 & 24.78 \\
FGSM-RS \textbf{+ RPS} & 51.09 & \textbf{36.75} & \textbf{37.18} & 64.95 & \textbf{39.18} & \textbf{38.36} \\ \hline
PGD-7 & 56.31 & 27.97 & 27.77 & 60.36 & 31.06 & 30.86 \\
PGD-7 \textbf{+ RPS} & 56.2 & \textbf{41.74} & \textbf{42.1} & 58.41 & \textbf{40.45} & \textbf{40.5} \\  \hline
\end{tabular}
}
\label{tab:sota_cifar100}
\vspace{-1em}
\end{table}

\begin{table}[t]
\centering
\caption{Evaluating RPS on two networks trained with FGSM-RS~\cite{wong2019fast} and PGD-7~\cite{madry2017towards} on the SVHN dataset.}
\vspace{-1em}
\resizebox{0.48\textwidth}{!}{
\begin{tabular}{ccccccc}
\toprule
\multicolumn{1}{l}{\textbf{}} & \multicolumn{3}{c}{\textbf{PreActResNet-18}} & \multicolumn{3}{c}{\textbf{WideResNet-32}} \\ \hline
\textbf{\begin{tabular}[c]{@{}c@{}}Adversarial \\ Training Method\end{tabular}} & \textbf{\begin{tabular}[c]{@{}c@{}}Natural \\ (\%)\end{tabular}} & \textbf{\begin{tabular}[c]{@{}c@{}}PGD-20 \\ (\%)\end{tabular}} & \textbf{\begin{tabular}[c]{@{}c@{}}PGD-100 \\ (\%)\end{tabular}} & \textbf{\begin{tabular}[c]{@{}c@{}}Natural \\ (\%)\end{tabular}} & \textbf{\begin{tabular}[c]{@{}c@{}}PGD-20 \\ (\%)\end{tabular}} & \textbf{\begin{tabular}[c]{@{}c@{}}PGD-100 \\ (\%)\end{tabular}}  \\ \hline \hline
FGSM-RS & 88.68 & 44.62 & 43.59 & 92.20 & 42.62 & 40.88  \\
FGSM-RS \textbf{+ RPS} & 86.46 & \textbf{53.51} & \textbf{53.92} & 93.83 & \textbf{57.99} & \textbf{57.42} \\ \hline \hline
PGD-7 & 86.81 & 51.53 & 50.98 & 91.07 & 54.39 & 53.53  \\
PGD-7 \textbf{+ RPS} & 87.22 & \textbf{61.84} & \textbf{61.64} & 91.05 & \textbf{65.59} & \textbf{64.62} \\ \bottomrule
\end{tabular}
}
\label{tab:svhn}
\vspace{-1em}
\end{table}

\begin{table}[t]
\centering
\caption{Evaluating RPS on top of two adversarial training methods (FGSM-RS~\cite{wong2019fast} and Free~\cite{shafahi2019adversarial}) on ResNet-50 under PGD-10 and PGD-50 attacks with $\epsilon=4$ on ImageNet.}
\vspace{-0.5em}
\resizebox{0.4\textwidth}{!}{
\begin{tabular}{cccc}
\toprule
\textbf{\begin{tabular}[c]{@{}c@{}}Adversarial \\ Training Method \end{tabular}} & \textbf{\begin{tabular}[c]{@{}c@{}}Natural \\ (\%)\end{tabular}} & \textbf{\begin{tabular}[c]{@{}c@{}}PGD-10 \\ (\%)\end{tabular}} & \textbf{\begin{tabular}[c]{@{}c@{}}PGD-50 \\ (\%)\end{tabular}}  \\\hline \hline
FGSM-RS & 55.45 & 30.28 & 30.18 \\
FGSM-RS \textbf{+ RPS} & \textbf{63.21} & \textbf{37.93} & \textbf{37.12}  \\ \hline
Free & 60.21 & 32.77 & 31.88   \\
Free \textbf{+ RPS} & \textbf{64.58} & \textbf{42.88} & \textbf{42.72} \\ \hline
\end{tabular}
}
\label{tab:imagenet}
\end{table}

\subsubsection{Benchmark on CIFAR-10/CIFAR-100/SVHN/ImageNet} 
\label{sec:exp_dataset}

\textbf{\\ \indent Benchmark on CIFAR-10.} As summarized in Tab.~\ref{tab:sota_cifar10}, we can observe that (1) RPS consistently enhances the robust accuracy under PGD attacks, largely outperforming SOTA adversarial training methods with a full precision. In particular, RPS achieves a 13.98\%/12.14\% higher robust accuracy under PGD-20 attacks on PreActResNet-18 and WideResNet-32, respectively, while notably improving the efficiency thanks to the low precision execution as evaluated in Sec.~\ref{sec:exp_accelerator}; and (2) RPS also enhances the robust accuracy by 13.57\%$\sim$22.60\% under PGD-20 attacks on top of FGSM/FGSM-RS. 

\textbf{Benchmark on CIFAR-100.} The observations on CIFAR-100 are consistent with those on CIFAR-10. In particular, RPS achieves a 10.61\%/13.77\% and 13.83\%/9.39\% higher robust accuracy on top of FGSM-RS/PGD-7 training under PGD-20 attacks on PreActResNet-18 and WideResNet-32, respectively.

\textbf{Benchmark on SVHN.} 
As shown in Tab.~\ref{tab:svhn}, RPS achieves a 8.89\%$\sim$15.37\% and 10.31\%$\sim$11.20\% higher robust accuracy under PGD-20 attacks and a comparable natural accuracy on top of FGSM-RS and PGD-7 training, respectively, indicating that RPS is generally effective on various tasks.

\textbf{Benchmark on ImageNet.} We further evaluate RPS on a larger scale dataset, i.e, ImageNet, as shown in Tab.~\ref{tab:imagenet}. We can observe that RPS achieves a \textbf{triple-win} in terms of the natural accuracy, robust accuracy, and model efficiency on top of both adversarial training methods. In particular, RPS achieves a 7.65\%/10.11\% higher robust accuracy over FGSM-RS~\cite{wong2019fast} and Free~\cite{shafahi2019adversarial}, respectively, under the PGD-10 attack, indicating our RPS's scalability and applicability on large-scale and complex datasets.

\begin{table}[t]
\centering
\caption{Evaluating RPS on two networks trained by PGD-7 under more strong attacks with $\epsilon$=8 and 12 on CIFAR-10.}
\vspace{-1em}
\centering
\resizebox{0.48\textwidth}{!}{
\begin{tabular}{ccccc}
\hline
\textbf{} & \multicolumn{2}{c}{\textbf{PreActResNet-18}} & \multicolumn{2}{c}{\textbf{WideResNet-32}} \\ \hline
Attack Type & PGD-7 & PGD-7 \textbf{+ RPS} & PGD-7 & PGD-7 \textbf{+ RPS} \\ \hline
\begin{tabular}[c]{@{}c@{}}AutoAttack ($\epsilon$=8)\end{tabular} & 47.18 & \textbf{54.56} & 51.66 & \textbf{58.54} \\
\begin{tabular}[c]{@{}c@{}}AutoAttack ($\epsilon$=12)\end{tabular} & 27.59 & \textbf{35.83} & 30.71 & \textbf{39.83} \\ \hline \hline
\begin{tabular}[c]{@{}c@{}}CW-Inf ($\epsilon$=8)\end{tabular} & 57.88 & \textbf{71.44} & 62.13 & \textbf{72.10} \\
\begin{tabular}[c]{@{}c@{}}CW-Inf ($\epsilon$=12)\end{tabular} & 46.70 & \textbf{65.57} & 50.14 & \textbf{66.99} \\ \hline \hline
\begin{tabular}[c]{@{}c@{}}Bandits ($\epsilon$=8)\end{tabular} & 59.75 & \textbf{71.75} & 63.49 & \textbf{68.50} \\
\begin{tabular}[c]{@{}c@{}}Bandits ($\epsilon$=12)\end{tabular} & 46.04 & \textbf{70.52} & 49.77 & \textbf{67.01} \\ \hline
\end{tabular}
}
\label{tab:strong_attack}
\vspace{-1em}
\end{table}

\subsubsection{Benchmark under more attacks}
\label{sec:exp_strong}
\textbf{\\} Considering many defense methods are found to be ineffective under stronger attacks, we evaluate RPS against more attack methods with different permutation strengths. As observed from Tab.~\ref{tab:strong_attack}, RPS consistently improves the robust accuracy across different attacks/models/distortions, e.g., a higher robust accuracy of 6.88\%$\sim$9.12\% under Auto-Attack, which is currently one of the strongest adaptive attacks, and more surprisingly, 9.97\%$\sim$18.87\% under the CW-Inf attack, where we find the poor transferability between different attack/inference precisions is more notable. In addition, RPS achieves a 5.01\%$\sim$24.48\% higher robustness accuracy under the Bandits attack which is a gradient-free attack, indicating that RPS does not suffer from the obfuscated gradient problem~\cite{athalye2018obfuscated}. In fact, we find RPS does not show any characteristic behavior for obfuscated gradients discussed in~\cite{athalye2018obfuscated}.

\begin{table}[t]
\centering
\caption{Evaluating RPS against the customized adaptive attack E-PGD on top of PreActResNet-18 on CIFAR-10/100.}
\vspace{-1em}
\centering
\resizebox{0.48\textwidth}{!}{
\begin{tabular}{ccccccc}
\hline
\textbf{} & \multicolumn{3}{c}{\textbf{CIFAR-10}} & \multicolumn{3}{c}{\textbf{CIFAR-100}} \\ \hline
\textbf{\begin{tabular}[c]{@{}c@{}}Adversarial\\ Training Method\end{tabular}} & \textbf{\begin{tabular}[c]{@{}c@{}}Natural \\ (\%)\end{tabular}} & \textbf{\begin{tabular}[c]{@{}c@{}}E-PGD\\ -20 (\%)\end{tabular}} & \textbf{\begin{tabular}[c]{@{}c@{}}E-PGD\\ -100 (\%)\end{tabular}} & \textbf{\begin{tabular}[c]{@{}c@{}}Natural \\ (\%)\end{tabular}} & \textbf{\begin{tabular}[c]{@{}c@{}}E-PGD\\ -20 (\%)\end{tabular}} & \textbf{\begin{tabular}[c]{@{}c@{}}E-PGD\\ -100 (\%)\end{tabular}} \\ \hline
PGD-7 & 82.02 & 51.17 & 50.93 & 56.31 & 27.97 & 27.77 \\
PGD-7 \textbf{+ RPS} & 82.16 & \textbf{60.14} & \textbf{60.08} & 56.23 & \textbf{37.58} & \textbf{37.89} \\ \hline
\end{tabular}
}
\label{tab:adaptive_attack}  
\vspace{-0.2em}
\end{table}

\subsubsection{Benchmark under adaptive attacks}
\textbf{\\} We further evaluate RPS via customizing an adaptive attack~\cite{tramer2020adaptive}, dubbed E-PGD, which generates perturbations based on the ensemble (i.e., the averaged ouptput) of all candidate precisions to make the attacks aware of all precisions, assuming that the adversaries know the adopted precision set in advance. As shown in Tab.~\ref{tab:adaptive_attack}, RPS still achieves a more than 8.97\% and 9.61\% higher robust accuracy over PGD-7 training on CIFAR-10 and CIFAR-100, respectively, indicating the consistent effectiveness of RPS.




\vspace{-0.5em}
\subsection{Evaluate 2-in-1 Accelerator’s architecture}
\label{sec:exp_accelerator}

\subsubsection{Benchmark with Bit Fusion and Stripes} 
\label{sec:exp_sota_precision_scalable}

\begin{figure}
    \centering
    \includegraphics[width=\linewidth]{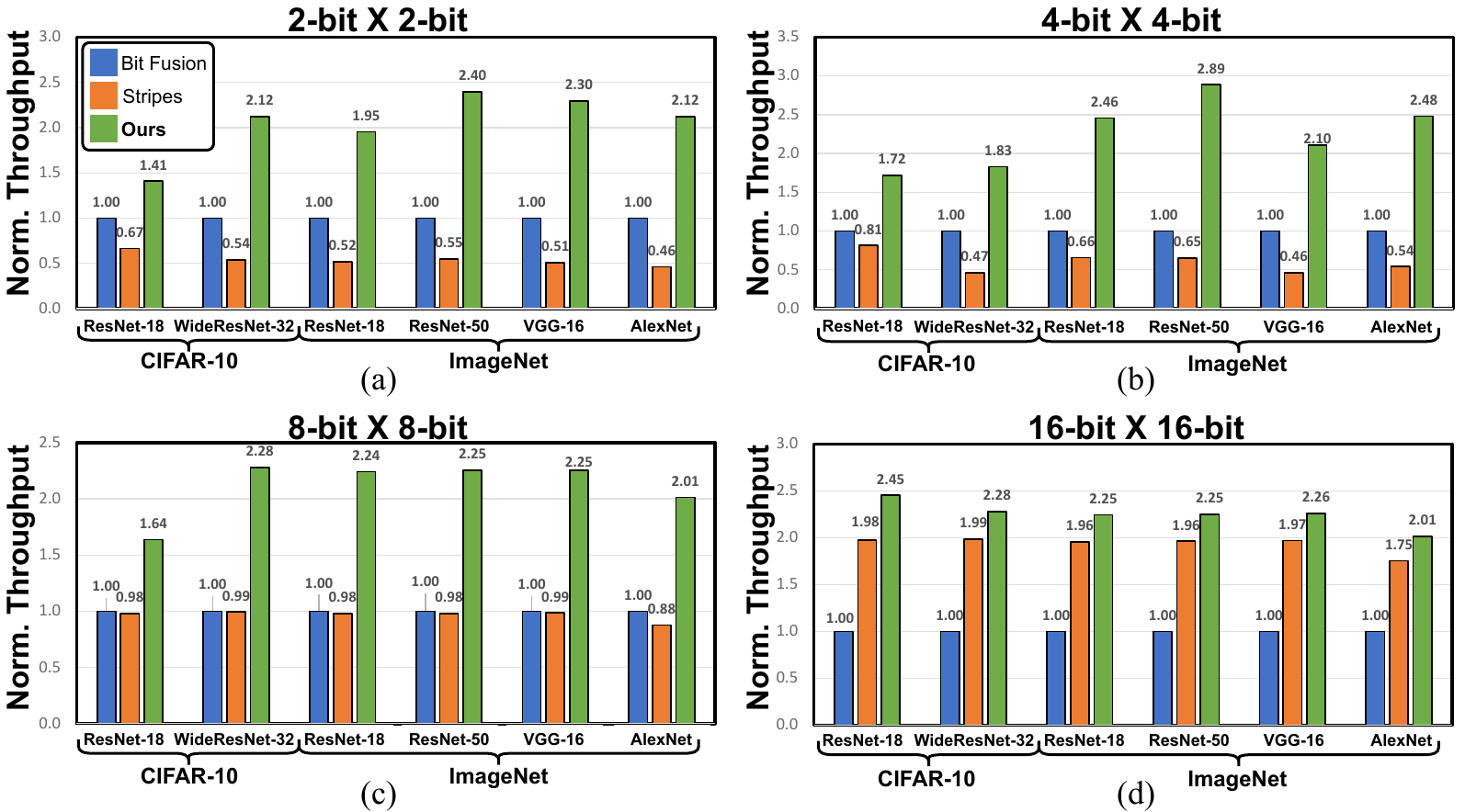}
    \vspace{-1.5em}
    \caption{Normalized throughput comparison among Bit Fusion, Stripes, and our \textit{2-in-1 Accelerator} on top of six networks and four execution precisions.}
    \label{fig:throughput}
\end{figure}

\textbf{\\ \indent Throughput comparison.}
We compare the throughput of Bit Fusion, Stripes, and our \textit{2-in-1 Accelerator} on top of six networks and four execution precisions in Fig.~\ref{fig:throughput}. All the throughput results are normalized to that of Bit Fusion. We can observe that our design outperforms the baselines across all the networks and precisions with a 1.41$\times$ $\sim$ 2.88$\times$ and 1.15$\times$ $\sim$ 4.59$\times$ higher throughput over Bit Fusion and Stripes, respectively.
Such improvement mainly comes from \underline{(1)} the high throughput/area of our proposed MAC unit architecture, and \underline{(2)} the effectiveness of our automated optimizer in reducing the memory stalls to fully utilize the capability of our MAC unit. For example, when using ResNet-50 on ImageNet with 4x4-bit, our MAC unit design boosts the throughput by 2.25$\times$ over Bit-Fusion, and our automated optimizer further improves the throughput by 1.28$\times$ via reducing the memory stalls.
In addition, we can observe that Bit Fusion shows a better throughput over Stripes for execution precisions lower than 8-bit while showing an inferior throughput at 16-bit, which is consistent with the analysis in Sec.~\ref{sec:bottleneck_trade-off} showing that Bit Fusion requires to execute each MAC unit four times for execution precisions higher than 8-bit. Although our accelerator adopts a similar manner to deal with 16-bit, it can still achieve a 1.15$\times$ higher throughput over Stripes, validating the superiority of the proposed spatial-temporal MAC design.

\begin{figure}
    \centering
    \includegraphics[width=\linewidth]{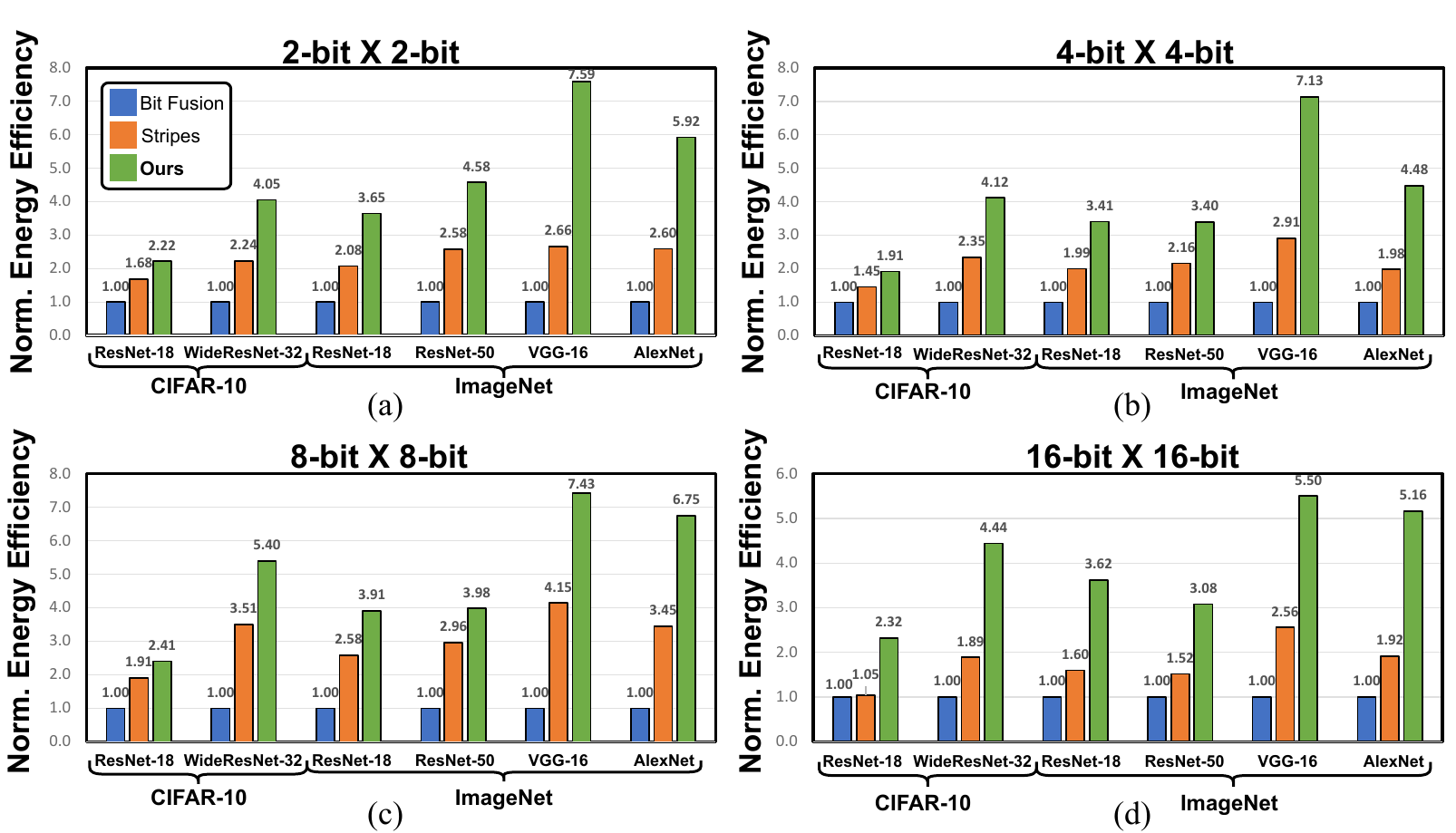}
    \vspace{-2em}
    \caption{Normalized energy efficiency comparison among Bit Fusion, Stripes, and our \textit{2-in-1 Accelerator} on top of six networks and four execution precisions.}
    \vspace{-1.5em}
    \label{fig:energy}
\end{figure}

\textbf{Energy efficiency comparison.}
We compare the energy efficiency of Bit Fusion, Stripes, and our \textit{2-in-1 Accelerator} on top of six networks and four execution precisions in Fig.~\ref{fig:energy}. All the energy efficiency results are normalized to that of Bit Fusion. We can observe that our proposed architecture consistently achieves the best energy efficiency across all the networks and precisions with a 1.91$\times$ $\sim$ 7.58$\times$ and 1.25$\times$ $\sim$ 2.85$\times$ energy efficiency over Bit Fusion and Stripes, respectively. Here we fully optimize the dataflow of Stripes so that it also outperforms Bit Fusion in energy efficiency.

\begin{figure}
    \centering
    \includegraphics[width=\linewidth]{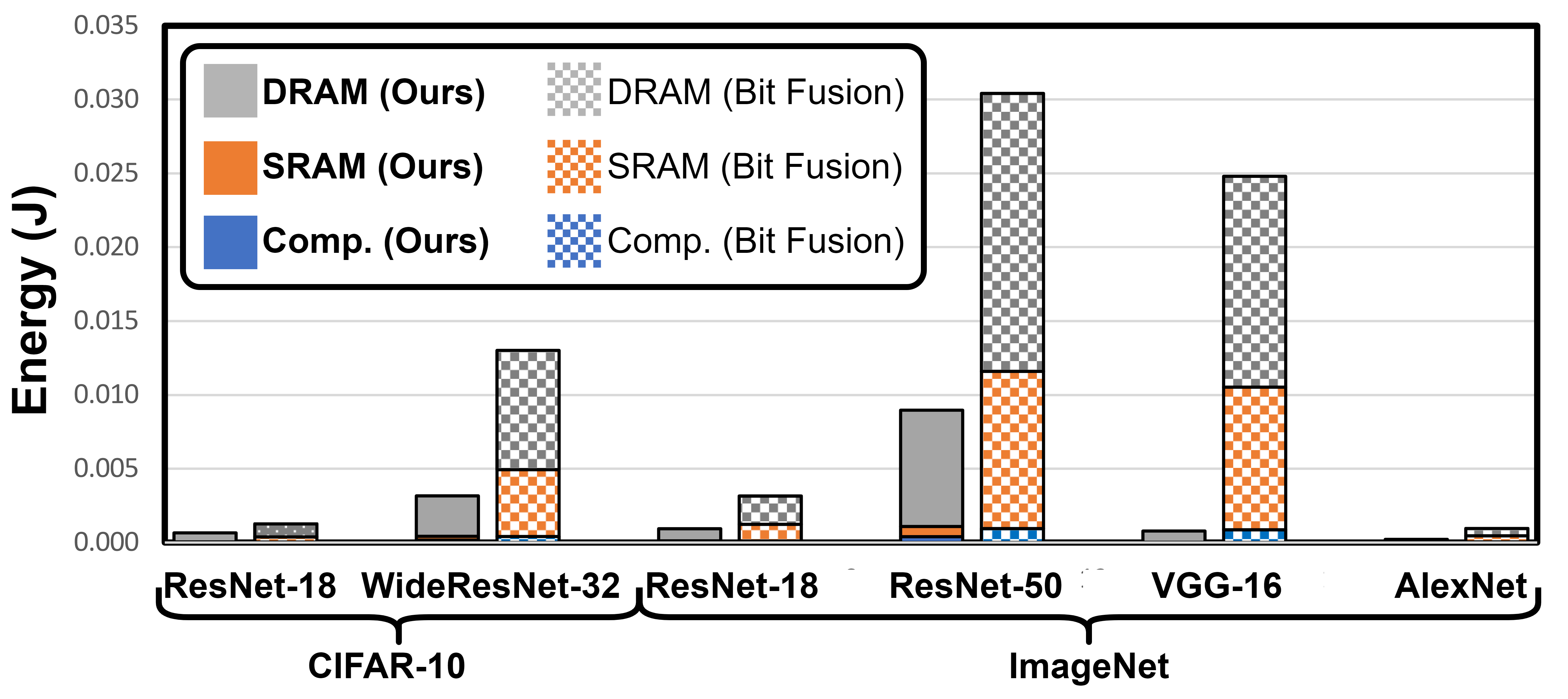}
    \vspace{-2em}
    \caption{Energy breakdown of our \textit{2-in-1 Accelerator} and Bit Fusion on six networks executed with 4-bit$\times$4-bit.}
    \label{fig:energy_breakdown}
\end{figure}

We also compare the energy breakdown between our design and Bit Fusion in Fig.~\ref{fig:energy_breakdown}. We can observe that although DRAM access still dominates the total energy, the energy costs for both the MAC computations and data movement (i.e., access DRAM and SRAM) are all reduced over Bit Fusion. The former is due to the higher energy efficiency/operation of our MAC unit and the latter is due to \underline{(1)} the new opportunities for better mapping strategies brought by the proposed MAC unit with better throughput/area and output reuses, and \underline{(2)} the effectiveness of our automated optimizer on a more flexible dataflow search space.

\begin{figure}
    \centering
    \includegraphics[width=\linewidth]{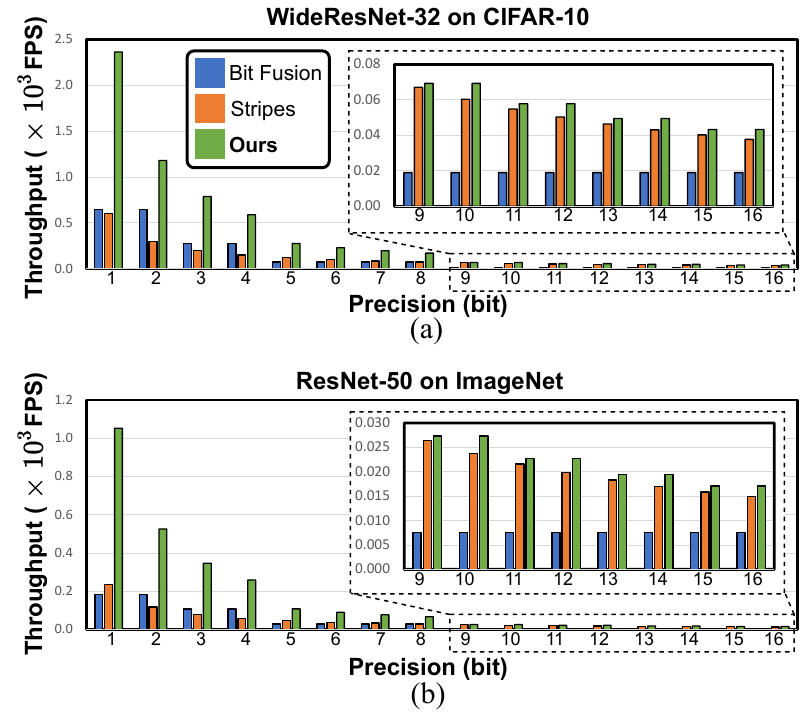}
    \vspace{-2em}
    \caption{Throughput under different precisions of Bit Fusion, Stripes, and our \textit{2-in-1 Accelerator} for accelerating WideResNet-32 on CIFAR-10 and ResNet-50 on ImageNet.}
    \label{fig:exp_throughput_bit}
\end{figure}

\textbf{Throughput evolution with the execution precision.}
To further validate the scalability along different execution precisions of our \textit{2-in-1 Accelerator} over spatial/temporal designs, we evaluate the throughput under different precisions (the same weight/input precision) of Bit Fusion, Stripes, and our design, when accelerating WideResNet-32 on CIFAR-10 and ResNet-50 on ImageNet. As observed in Fig.~\ref{fig:exp_throughput_bit}, our \textit{2-in-1 Accelerator} shows both superior efficiency and flexibility as it \underline{(1)} consistently outperforms both the baselines under all the precisions with up to 4.42$\times$ higher throughput, and \underline{(2)} achieves a consistent improvement in the throughput as the precision decreases. In addition, under execution precisions lower than 8-bit which are the common choices of recent quantization works~\cite{jung2019learning, bhalgat2020lsq+, esser2019learned, park2020profit}, our design shows more than 1.82$\times$ higher throughput compared with the best baseline; and under execution precisions higher than 8-bit which are inferior choices for spatial designs as analyzed in Sec.~\ref{sec:bottleneck_trade-off}, our design still achieves a higher throughput over Stripes, which benefits from the spatial-temporal design of our MAC unit.


\subsubsection{Benchmark with robustness-aware accelerators} 
\label{sec:exp_sota_robustness}
\textbf{\\} Boosting both robustness and efficiency in one accelerator is a significant feature and benefit of our \textit{2-in-1 Accelerator}. We further benchmark with a SOTA robustness-aware accelerator DNNGuard~\cite{wang2020dnnguard} to show the superiority of our framework. In particular, we compare the throughput/area of our \textit{2-in-1 Accelerator} and that of DNNGuard on AlexNet, VGG-16, and ResNet-50 which are reported by~\cite{wang2020dnnguard}. We find that our design achieves a 36.5$\times$/17.9$\times$, 19.3$\times$/9.5$\times$, and 12.8$\times$/6.4$\times$ higher throughput compared with DNNGuard when adopting 4$\sim$8-bit/4$\sim$16-bit for accelerating AlexNet, VGG-16, and ResNet-50, respectively. This indicates the superiority and practicality of deploying our \textit{2-in-1 Accelerator} in real-world IoT applications where both security and efficiency matter.

\begin{figure}[h]
    \centering
    \vspace{-0.5em}
    \includegraphics[width=0.65\linewidth]{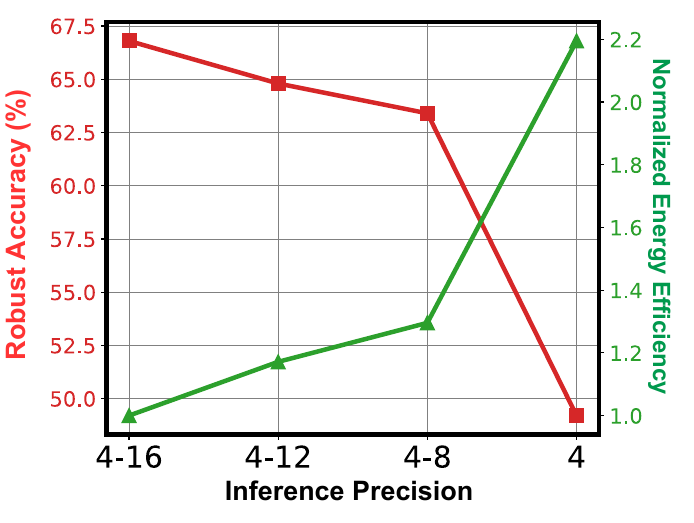}
    \vspace{-1em}
    \caption{\textit{2-in-1 Accelerator}'s instant robustness-efficiency trade-off on top of WideResNet-32 and CIFAR-10.}
    \vspace{-0.5em}
    \label{fig:exp_instant}
\end{figure}

\vspace{-0.5em}
\subsection{Instant robustness-efficiency trade-offs of the 2-in-1 Accelerator}
\label{sec:exp_instant_trade_off}
As analyzed in Sec.~\ref{sec:instant}, our \textit{2-in-1 Accelerator} also features the capability to enable instant robustness-efficiency trade-offs at run-time to adapt to both the safety conditions of the environments and the remaining power on the device. We show an example of executing WideResNet-32 with CIFAR-10 inputs on our \textit{2-in-1 Accelerator} with different execution precisions (RPS with 4$\sim$16-bit, 4$\sim$12-bit ,4$\sim$8-bit, and static 4-bit) and record the robust accuracy and the (averaged) energy efficiency. As shown in Fig.~\ref{fig:exp_instant}, our \textit{2-in-1 Accelerator} can instantly switch between high precision sets, low precision sets, and static low precision to balance the achieved robustness and efficiency with a comparable natural accuracy (within 81.5\%$\sim$84.7\%).

\section{Related Works}

\textbf{\indent Adversarial attacks and defenses.}
\cite{goodfellow2014explaining} shows that small permutations onto the inputs can mislead DNNs' decisions, which is known as adversarial attacks. 
Later, stronger attacks, including both white-box~\cite{madry2017towards, croce2020reliable, carlini2017towards, papernot2016limitations, moosavi2016deepfool} and black-box ones~\cite{chen2017zoo, ilyas2018prior, andriushchenko2020square, guo2019simple, ilyas2018black}, are proposed to aggressively degrade the accuracy of the target DNN models.
To defend DNNs against adversarial attacks, adversarial training~\cite{shafahi2019adversarial,madry2017towards,wong2019fast,tramer2017ensemble}, which augments the training set with adversarial samples generated during training, is currently the most effective method. In parallel, other defense methods~\cite{guo2017countering, buckman2018thermometer,song2017pixeldefend,xu2017feature,liao2018defense,metzen2017detecting,feinman2017detecting,li2018certified,wu2020adversarial} have also been proposed.
There has been a continuous competition between adversaries and defenders, and the readers are referred to~\cite{akhtar2018threat, chakraborty2018adversarial} for more discussions.

\textbf{Robustness of quantized models.}
As both robustness and efficiency are critical for most DNN applications, pioneering works have strived to design robust quantized DNNs. In particular,~\cite{galloway2017attacking, panda2019discretization} propose robust binary neural networks (BNNs) and~\cite{rakin2018defend} adopts tanh-based quantization to increase robustness, while these works have been observed to suffer from the obfuscated gradient problem~\cite{athalye2018obfuscated, papernot2017practical}, which is a false sense of security. Later,~\cite{lin2019defensive} finds that quantized DNNs are actually more vulnerable to adversarial attacks due to the error amplification effect, i.e., the magnitude of adversarial perturbation is amplified when passing through the DNN layers.
To tackle this effect,~\cite{lin2019defensive, shkolnik2020robust} propose robustness-aware regularization methods for DNN training, and~\cite{song2020improving} retrains the network via feedback learning~\cite{song2019feedback}. In addition,~\cite{panda2020quanos} searches for layerwise precision and~\cite{gui2019model} constructs a unified formulation to balance and enforce the models' robustness and compactness.
In contrast, our RPS algorithm \textit{leverages quantization} to aggressively enhance robustness, which even largely surpasses the full-precision models.


\textbf{Precision-scalable accelerators.}
To support variable precisions for different DNN models/layers, various precision-scalable accelerators have been proposed to dynamically and flexibly handle the 
varied workloads, which can be categorized into two classes, i.e., \textit{temporal} and \textit{spatial} designs. For temporal designs, pioneering works, such as Stripes~\cite{judd2016stripes}, LOOM~\cite{sharify2018loom}, and Tartan~\cite{delmas2017tartan}, adopt bit-serial MAC units to provide precision configurability, which can flexibly handle any prevision yet suffer from inferior efficiency per area over their spatial counterparts~\cite{sharma2018bit, camus2019review}, and more recently UNPU~\cite{lee2018unpu} fabricates a bit-serial DNN accelerator to support variable weight precisions while the activations use full precision. For spatial designs, Bit Fusion~\cite{sharma2018bit} proposes to use combinational logic to dynamically compose and decompose 2-bit multipliers to construct variable-precision MAC units; Later, BitBlade~\cite{ryu2019bitblade} improves Bit Fusion via pulling out the shifting logic of each MAC unit and sharing it across the multipliers to reduce the area overhead; 
DVAFS~\cite{moons2017dvafs, moons20160} propose to turn off parts of the multipliers at low precision to increase the energy efficiency at a constant throughput; and DeepRecon~\cite{rzayev2017deeprecon} skips parts of the pipeline stages of a floating-point-multiplier to support either one 16-bit, two 12-bit, or four 8-bit multiplications. 
Detailed benchmarks for different precision-scalable MAC unit architectures can be found in~\cite{camus2019review}. Our proposed MAC unit architecture marries the best of both temporal and spatial designs and is integrated to construct a new precision-scalable accelerator, which consistently outperforms SOTA designs under various settings. 
\textbf{Robustness-aware DNN accelerators.}
Despite their importance for real-world applications, the art of robustness-aware DNN accelerators is still in its infancy. Pioneering works~\cite{wang2020dnnguard, rouhani2018deepfense, gan2020ptolemy} aim to defend against adversarial attacks within DNN accelerators at a cost of additional detection networks/modules. In particular, \cite{rouhani2018deepfense} proposes an end-to-end framework based on the voting results of multiple detectors, in parallel with the execution of the target DNN to detect malicious inputs during inference;
\cite{wang2020dnnguard} proposes an elastic heterogeneous DNN accelerator architecture to orchestrate the simultaneous execution of the target DNN and the detection network for detecting adversarial samples via an elastic management of the on-chip buffer and PE computing resources; \cite{gan2020ptolemy} builds an algorithm-architecture co-designed system to detect adversarial attacks during inference via a random forest module applied on top of the extracted features from the run-time activations. In addition, \cite{qin2020design} builds a robustness-aware accelerator based on BNNs which, however, suffers from the obfuscated gradient problem~\cite{athalye2018obfuscated} and~\cite{guo2019hardware} strives to speed up the attack generation instead of the defense. Nevertheless, all the existing defensive accelerators rely on additional detection networks/modules to detect adversarial samples at inference time, and thus inevitably introduce additional energy/latency/area overheads that compromise efficiency. In contrast, our work exploits the robustness within a DNN model via the proposed RPS algorithm to win both robustness and efficiency within one accelerator without introducing any extra modules.

\vspace{-1em}
\section{Conclusion}

Existing DNN accelerators mostly tackle only either efficiency or adversarial robustness while neglecting or even sacrificing the other. In this work, we propose the \textit{2-in-1 Accelerator}, aiming at winning both the adversarial robustness and efficiency of DNN accelerators. \textit{2-in-1 Accelerator} integrates a Random Precision Switch (RPS) algorithm that can effectively defend DNNs against adversarial attacks and a new precision-scalable accelerator featuring a spatial-temporal MAC unit architecture to boost both the achievable efficiency and flexibility and a systematically optimized dataflow generated by our generic accelerator optimizer. Extensive experiments and ablation studies validate our \textit{2-in-1 Accelerator}'s effectiveness and we believe our \textit{2-in-1 Accelerator} has opened up a new perspective for designing robust and efficient accelerators.

\begin{acks}

The work is supported by the NSF RTML program (Award number: 1937592), the NSF NeTS program (Award number: 1801865), and the NSF MLWiNS program (Award number: 2003137).

\end{acks}

\bibliographystyle{ACM-Reference-Format}
\bibliography{ref}

\end{document}